\documentclass{article}

\usepackage{arxiv}

\usepackage{amsmath,amsfonts,amsthm,color}
\usepackage{graphicx} % 引入graphicx包，用于插入图片
\usepackage{subcaption} % 引入subcaption包，用于创建子图
\usepackage{booktabs} % 用于表格中的 \toprule, \midrule, \bottomrule
\usepackage{multirow} % 用于表格中的 \multirow
\usepackage{amssymb}  % 用于插入符号 \ding{55}
\usepackage{pifont}

\usepackage[utf8]{inputenc} % allow utf-8 input
\usepackage[T1]{fontenc}    % use 8-bit T1 fonts
\usepackage{hyperref}       % hyperlinks
\usepackage{url}            % simple URL typesetting
\usepackage{booktabs}       % professional-quality tables
\usepackage{amsfonts}       % blackboard math symbols
\usepackage{nicefrac}       % compact symbols for 1/2, etc.
\usepackage{microtype}      % microtypography
\usepackage{cleveref}       % smart cross-referencing
\usepackage{lipsum}         % Can be removed after putting your text content
\usepackage{graphicx}
\usepackage{natbib}
\usepackage{doi}

\title{Brain-Inspired Online Adaptation for Remote Sensing with Spiking Neural Network}

\newif\ifuniqueAffiliation
\uniqueAffiliationtrue
\ifuniqueAffiliation 
\author{
  Dexin Duan\thanks{Brain-Inspired Application Technology Center (BATC), School of Electronic Information and Electrical Engineering, Shanghai Jiao Tong University, Shanghai, China, 200240.} \\
  \And
  \hspace{1mm}Peilin Liu\footnotemark[1] \\
  \And
  \hspace{1mm}Bingwei Hui\thanks{ATR Key Lab, School of Electronic Science and Engineering, National University of Defense Technology, Changsha, China 410073.} \\
  \And
  \hspace{1mm}Fei Wen\footnotemark[1] \thanks{Corresponding Author: wenfei@sjtu.edu.cn}\\
}
\hypersetup{
pdftitle={Brain-Inspired Online Adaptation for Remote Sensing with Spiking Neural Network},
pdfsubject={q-bio.NC, q-bio.QM},
pdfauthor={David S.~Hippocampus, Elias D.~Striatum},
pdfkeywords={First keyword, Second keyword, More},
}

\begin{document}
\maketitle

\begin{abstract}
On-device computing, or edge computing, is becoming increasingly important for remote sensing, particularly in applications like deep network-based perception on on-orbit satellites and unmanned aerial vehicles (UAVs). In these scenarios, two brain-like capabilities are crucial for remote sensing models: \textit{(1) high energy efficiency}, allowing the model to operate on edge devices with limited computing resources, and \textit{(2) online adaptation}, enabling the model to quickly adapt to environmental variations, weather changes, and sensor drift.
This work addresses these needs by proposing an online adaptation framework based on spiking neural networks (SNNs) for remote sensing. Starting with a pretrained SNN model, we design an efficient, unsupervised online adaptation algorithm,
which adopts an approximation of the BPTT algorithm and only involves forward-in-time computation that significantly reduces the computational complexity of SNN adaptation learning. 
Besides, we propose an adaptive activation scaling scheme to boost online SNN adaptation performance, particularly in low time-steps. Furthermore, for the more challenging remote sensing detection task, we propose a confidence-based instance weighting scheme, which substantially improves adaptation performance in the detection task. To our knowledge, this work is the first to address the online adaptation of SNNs. Extensive experiments on seven benchmark datasets across classification, segmentation, and detection tasks demonstrate that our proposed method significantly outperforms existing domain adaptation and domain generalization approaches under varying weather conditions. The proposed method enables energy-efficient and fast online adaptation on edge devices, and has much potential in applications such as remote perception on on-orbit satellites and UAV. 
\end{abstract}

% keywords can be removed

\section{Introduction}
In the past a few years, benefited from the rapid development of deep learning techniques,  deep neural network based methods have demonstrated impressive performance in various remote sensing applications such as scene classification \cite{DLrs1}, \cite{DLrs2}, semantic segmentation \cite{DLrs3}, geological interpretation \cite{DLrsinterpretation}, image fusion \cite{fusion} and detection \cite{rsdetection1,rsdetection2}. However, the increasingly large models pose significant challenges for on-device processing on platforms such as on-orbit satellites. 
On-device computing is becoming more and more important for some remote sensing applications, 
e.g., perception processing of remote sensing images on on-orbit satellites and unmanned aerial vehicles (UAVs).
In these scenarios, two brain-like capabilities are desired for remote sensing models, \textit{1) high energy efficiency}, allowing the model to operate on edge devices with limited computing resource such as on-orbit satellites and UAVs, \textit{2) online adaptation}, enabling the model to fast adapt to environmental variations, weather changes and sensor drifts.

The human brain, as the most complex and mysterious natural information processing system in the world, is an ultra complex neuron system with about 100 billion neurons and more than 1000 trillion synapse connections\cite{brainneurons}. However, 
the human brain only consumes about 20 watts\cite{BrainPower}, which is several orders of magnitude more energy efficient than modern computers. The human brain works in an ultra efficient event-driven way, by using spikes to encode and transmit information between neurons.
Moreover, the human brain can fast adapt to diverse varying environments and has extraordinary online learning ability. 
 
Inspired by the brain,  
spiking neural networks (SNN) emulates the spiking encoding and processing mechanism to achieve high energy efficiency.
Recent years have witnessed significant progress in SNN research \cite{snnbig1, snnbig3, SLTT, Dspike, snnNMI1, online, parallelsnn, TET, spikingjelly, stbp} with applications in image recognition \cite{snnimage}, object detection \cite{spikingyolo}, segmentation \cite{spikingseg}, speech recognition\cite{snnspeech}, and time-series prediction\cite{snnseries}. Recently, some researchers have applied SNN in remote sensing processing.
For example, the work \cite{snnrs1} introduces an approximate derivative algorithm for SNN to extract the spatial-spectral features of HSIs images. The work \cite{snnrs2} introduces an attention mechanism to realize real-time SNN based classification of HSI images. The work \cite{snnrs3} proposes an efficient SNN  transformer for satellite on-orbit computing. 

On the other hand, 
for on-device processing in remote sensing, 
 e.g., on on-orbit satellites and UAVs,
it is desired for a pre-trained perception model that can well generalize or fast
adapt to environmental variations, weather changes, and/or sensor drift.
In the field of remote sensing, there exists a number of recent works that aim to improve the cross-domain performance of deep models. 
For instance, the works \cite{li26DA} and \cite{xiong28DA} propose novel methods for cross-source image retrieval based on source-invariant deep hashing CNNs and cycle GAN. 
The work \cite{ma27DA} proposes a strategy based on teacher-ensemble learning and knowledge distillation
for cross-source image retrieval. The work \cite{udars1} introduces a novel method that uses high-level feature alignment to narrow the difference between the source and target domains at the semantic level. 
Moreover, in \cite{rsda1} and \cite{rsda2}, adversarial domain adaptation methods 
have been proposed for pixel-level classification of very high resolution images and change detection, respectively. 

However, the above methods  rely heavily
on labeled samples or require multiple epochs 
of training or fine-tuning, making them impractical 
for real-world deployment on edge devices for fast online adaptation.
Meanwhile, as the models becoming larger and larger,
the high energy consumption becomes another major issue for deployment on edge devices.

To address the these issues, 
we propose an efficient online adaptation framework for remote sensing based on brain-inspired SNN. 
Starting with a pretrained SNN model, we design an efficient, unsupervised online adaptation algorithm. It approximates the backpropagation through time (BPTT) method by partially decoupling the temporal gradient to achieve forward-in-time optimization, which significantly reduces the computational complexity of SNN adaptation learning. Additionally, we introduce an adaptive activation scaling scheme to boost online adaptation performance of SNN, particularly in low time-steps. Furthermore, for the remote sensing detection task, we develop a confidence-based instance weighting scheme, which substantially improves adaptation performance on detection.

The main contributions are summarized as follows.
\begin{itemize}
\item[1)]
We propose an online adaptation framework for SNN based remote sensing. 
With a pretrained SNN model from ANN-SNN conversion, 
we design an efficient, unsupervised online adaptation algorithm.
It significantly reduces the computational complexity by approximating the BPTT algorithm to achieve forward-in-time optimization in the adaptation learning of the SNN model.

\item[2)]
We design an adaptive activation scaling scheme to dynamically control the activation scale of a SNN model in the adaptation learning procedure, which significantly enhances the adaptation performance, particularly in low time-steps. 

\item[3)]
We design a confidence-based instance weighting scheme to significantly improve adaptation performance in the more challenging detection task. This approach selects high-confidence instances for model updating, thereby mitigating the influence  of low-confidence instances.

\item[4)]
Extensive evaluation of the proposed method in comparison with
representative ANN adaptation methods has been
conducted on seven benchmarking datasets, 
including three classification datasets (RSSCN, WHU-RS19, and AID), 
two semantic segmentation datasets (DLRSD and WHDLD), 
and two detection datasets (RSOD and LEVIR).
The results demonstrate that the proposed method significantly 
outperforms existing domain adaptation and domain generalization 
approaches under varying weather conditions.
\end{itemize}

While online unsupervised adaptation have been extensively 
studied for ANN, to our knowledge this work is the first to address
the online adaptation of SNN.
As presented in this work, 
extending existing  adaptation methods developed for ANN
to SNN is not straightforward.
To achieve satisfactory adaptation performance for SNN,
we have proposed an adaptive activation scaling approach, 
a confidence-based instance weighting approach.
Besides, simplified online SNN optimization algorithm is employed to
achieve computationally efficient adaptation.
Moreover, to support diverse tasks including classification, segmentation, and detection, this work mainly considers source models from ANN-SNN conversion.
However, the proposed method can be straightforwardly 
applied to directly trained SNN models. 

The rest of this paper is organized as follows. Section 2
% \uppercase\expandafter{\romannumeral2} 
introduces the detailed components of our proposed SNN adaptation framework. Section 3
% \uppercase\expandafter{\romannumeral3} 
introduces the experiment setup, including the datasets and implementation details. Section 4
% \uppercase\expandafter{\romannumeral4} 
presents the results and analysis, including experiment results, ablation study, and energy efficiency analysis. Finally, Section 5
% \uppercase\expandafter{\romannumeral5} 
provides conclusions and future work.

\section{Methodology}
\label{sec:headings}

The proposed SNN adaptation framework aims to fast adapt to test data 
with high computational efficiency. %It is compatible with most existing SNN architectures. 
In this section, we present the its components and implementation details.

\begin{figure*}
\center{\includegraphics[width=16cm]{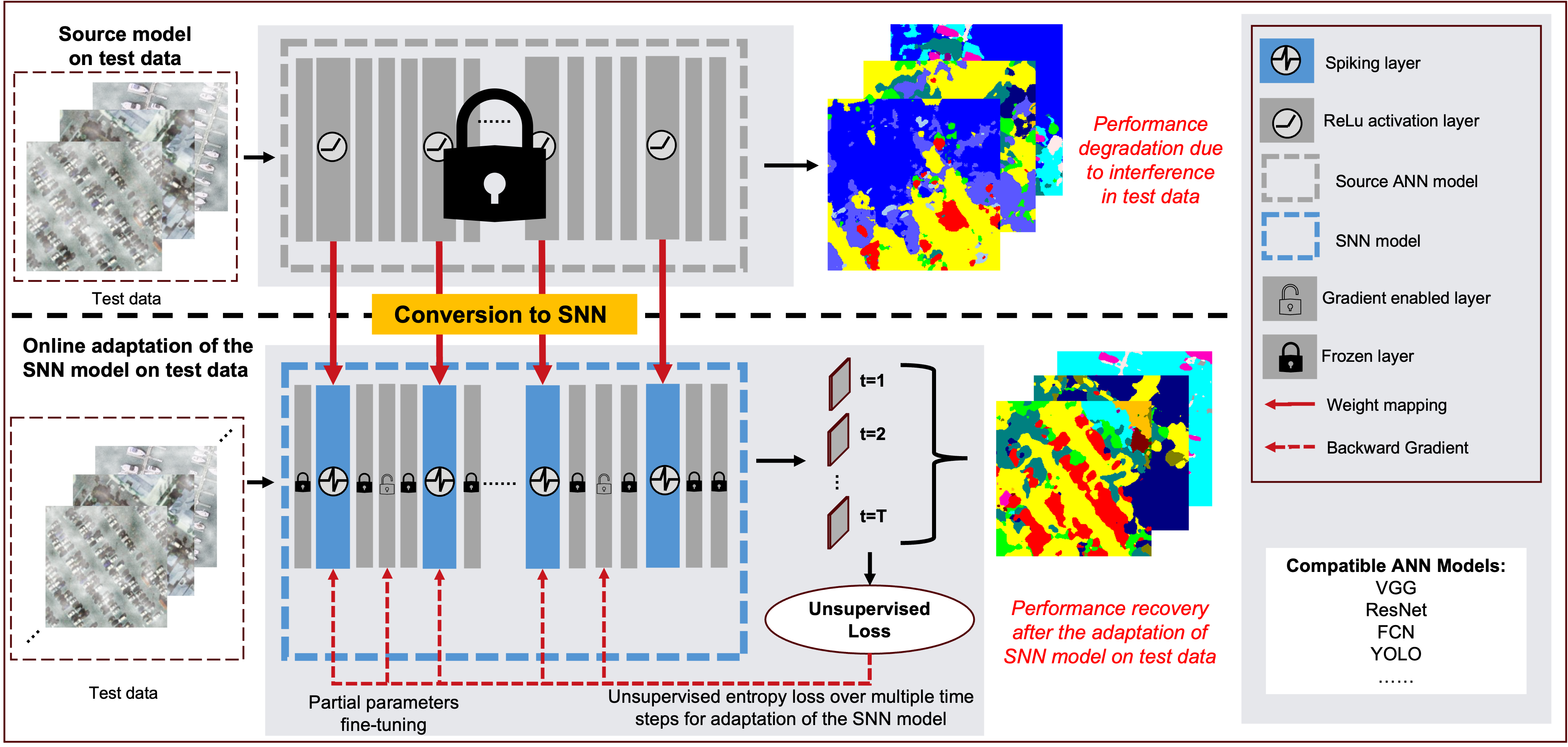}} 
\caption{The framework of proposed SNN adaptation pipeline.
A pre-trained source model may degrade significantly on test data in wild-world scenarios, e.g., diverse weather conditions.
The proposed method adapts a source SNN model on the test data in an online manner based on unsupervised loss, which can significantly improve the SNN model's performance in such scenarios. 
}
\label{pipeline_fig} 
\end{figure*}
\subsection{Spiking Neuron Model and Backpropagation Through Time}\label{IFBPTT}

Drawing inspiration from these biological neural networks, 
various spiking neuron models have been developed, 
such as the Hodgkin–Huxley model \cite{HHmodel}, 
the integrate-and-fire (IF) and leaky integrate-and-fire (LIF) models \cite{LIFneuron}, 
and the Izhikevich model \cite{izhikevich}. Among them, 
the IF model simplifies neuron dynamics while preserving essential characteristics and, most importantly, suitable for constructing
SNN models from ANN-to-SNN conversion as will described later.
The membrane potential $v$ of an IF model can be expressed as
\begin{equation}\label{lif1}
\frac{{d}v(t)}{{d}t}=v(t)+X(t),
\end{equation}
where $X(t)$ is the input current at time $t$.
By accumulating membrane potential over time, the IF neuron 
fires a spike when the membrane potential reaches a threshold $V_{th}$, 
and then is reset by substracting $V_{th}$. 
To facilitate numerical simulation, the dynamic model (\ref{lif1}) for a neuron $i$ is typically discretized  as
\begin{equation}\label{lif2}
\begin{cases}
v_i\left[t+1\right]=v_i[t]-V_{th}\cdot s_i[t]+\sum_jw_{ij}s_j[t]+b_i,\\
s_i[t+1]=H(v_i\left[t+1\right]-V_{th}),
\end{cases}
\end{equation}
where $s_i\in\{0,1\}$ represents the spike emission, $H(\cdot)$ stands for the Heaviside activation function, $w_{ij}$ denotes the weight between neuron $j$ and neuron $i$, and $b_i$ is the bias.

Considering the non-differentiable feature of the Heaviside function, to effectively train SNNs by backpropagation, surrogate gradient is widely adopted \cite{snnsg}, \cite{snnrs1}, \cite{snnrs2}. Using surrogate gradient, error signals for updating the weights can be propagated by backpropagation through time (BPTT) in both the spatial and temporal directions. 
%BPTT unfolds the iterative update equation in equation (\ref{lif1}) and back-propagates along the computational graph. 
For example, consider a multi-layer spiking network with IF neurons as
\begin{equation}\label{lif3}
\begin{cases}
\mathbf{v}^{l}[t+1]=\mathbf{v}^l[t]-V_{th}\cdot \mathbf{s}^{l}[t]+\mathbf{W}^{l-1}\mathbf{s}^{l-1}[t]+\mathbf{b}^l,\\
\mathbf{s}^l[t+1]=H\left(\mathbf{v}^l[t+1]-V_{th}\right),
\end{cases}
\end{equation}
where $\mathbf{v}^l\in\mathbb{R}^{n_l}$ collects the membrane potential of $n_l$ neurons in the $l$-th layer, $\mathbf{s}^l\in\{0,1\}^{n_l}$ denotes the emitted spikes by the neurons in the $l$-th layer, $\mathbf{W}^{l-1}\in\mathbb{R}^{n_l\times n_{l-1}}$ stands for the weights connecting the $(l-1)$-th and $l$-th layers.
Moreover, let $\mathcal{L}$ denote the loss function computed
based on the network output over $T$ time-steps.
Then, using BPTT to train the network, the gradient of the loss  $\mathcal{L}$ with respect to the weights $\mathbf{W}^l$ can be calculated as
\begin{equation}
\label{bptt} 
\begin{split}
&\frac{\partial \mathcal{L}}{\partial\mathbf{W}^l}=\sum_{t=1}^T\frac{\partial \mathcal{L}}{\partial\mathbf{s}^{l+1}[t]}{\frac{\partial\mathbf{s}^{l+1}[t]}{\partial\mathbf{v}^{l+1}[t]}}\left.\Bigg(\frac{\partial\mathbf{v}^{l+1}[t]}{\partial\mathbf{W}^l}+
\sum_{t^\prime<t}\prod_{k=t-1}^{t^\prime} 
\left(1+
\frac{\partial\mathbf{v}^{l+1}[k+1]}{\partial\mathbf{s}^{l+1}[k]}\frac{\partial\mathbf{s}^{l+1}[k]}{\partial\mathbf{v}^{l+1}[k]}
\right)
\frac{\partial\mathbf{v}^{l+1}[t^\prime]}{\partial\mathbf{W}^l}\right.\Bigg),
\end{split}
\end{equation}
where we used $\frac{\partial\mathbf{v}^{l+1}[k+1]}{\partial\mathbf{v}^{l+1}[k]}=1$ for IF neuron. The non-differentiable term $\frac{\partial\mathbf{s}^{l+1}[t]}{\partial\mathbf{v}^{l+1}[t]}$ is replaced by surrogate functions %\cite{snnsg} 
to approximately compute the backward errors.
Recently, surrogate gradient based methods have shown 
remarkable effectiveness in training SNNs \cite{snnsg,snnrs1}.
Though effective, 
the computational and memory costs of BPTT scale linearly 
with the number of time-steps $T$, as it requires unfolding 
the the backward computational graph over the $T$ time-steps.
 
%介绍权重拷贝映射以及ANN-SNN转换原理，以及在转换过程中引入的clip系数。
\subsection{ANN-SNN Conversion}\label{ANN-SNN}

In this work, we mainly consider pre-trained source models from
ANN-SNN conversion, as it facilitates to fully utilize existing 
ANN models developed for various remote sensing tasks, such as classification,
segmentation, and detection. 
The proposed framework is illustrated in Fig. 1. 
However, the proposed adaptation 
method can be straightforwardly applied to directly trained SNN models.

Before proceeding to the proposed online adaptation method,
we first briefly introduce ANN-SNN conversion \cite{ann2snn1,ann2snn2,ann2snn3,rmp,ann2snn4}. 
Consider the most widely used activation function ReLU \cite{RELU2},
the activation of neuron $i$ in layer $l$ can be expressed as
\begin{equation}
{a}_{l,i}=\max\left(\left(\boldsymbol{W}_{l-1,i}^{ANN}\right)^{\top}\boldsymbol{a}_{l-1}+{b}_{l,i}^{ANN},0\right),
\end{equation}
where $^{\top}$ stands for the transpose operation, 
$\boldsymbol{a}_{l-1}\in\mathbb{R}^m$ collects the activation of the $m$ neurons in layer $l-1$, $\boldsymbol{W}^{ANN}_{l-1,i}\in\mathbb{R}^m$ denotes the weights from the $(l-1)$-th layer to neuron $i$ in layer $l$,
and ${b}_{l-1,i}$ denotes the bias of neuron $i$ in layer $l$. 
Here, we use the superscript $^{ANN}$
to differentiate between the weights of ANN and that of the converted SNN.
By explicitly counting the range of activation values in the $l$-th layer
to determine the maximum activation $a^l_{max}$, 
the activation ${a}_{l,i}$ can be normalized as
\begin{equation}\label{relu2}
\overline{{a}}_{l,i}%=\frac{\boldsymbol{a}_l{a_{max}^l}
=clip\left(\frac{\left(\boldsymbol{W}_{l,i}^{ANN}\right)^{\top} \boldsymbol{a}_{l-1}}{a_{max}^l}+\frac{{b}_{l-1,i}^{ANN}}{a_{max}^l},0,1\right).
\end{equation}

For IF neurons, the accumulated voltage $v_{l,i}(T)$ of neuron $i$ in layer $l$ after $T$ time-steps can be expressed as the weighted membrane voltage minus the voltage drop incurred by the spike firing of the previous spiking layer at all times as
\begin{equation}\label{vt}
{v}_{l,i}(T)=\left(\boldsymbol{W}_{l-1,i}^{SNN}\right)^{\top}\sum_t\boldsymbol{s}_{l-1}(t)+\sum_t{b}_{l,i}^{SNN}-v_{th}\sum_t{s}_{l,i}(t),
\end{equation}
where $\boldsymbol{W}^{SNN}_{l-1,i}\in\mathbb{R}^m$ denotes 
the weights connecting from the $(l-1)$-th layer to neuron $i$ 
in layer $l$ in the SNN network, 
$\boldsymbol{s}_{l-1}(t)\in\mathbb{R}^m$ denotes 
the spike emission of layer $(l-1)$ at time-step $t$, 
${s}_{l,i}(t)$ is the spike emission of neuron $i$ 
in layer $l$ at time-step $t$,
$v_{th}$ is the firing threshold. 
Rewriting equation (\ref{vt}) as a form of firing rate
$r_{l,i}(T)=\frac{\Sigma_ts_{l,i}(t)}T$, it follows that
\begin{equation}
\label{vt_fr}
r_{l,i}(T)=\frac{\left(\boldsymbol{W}_{l-1,i}^{SNN}\right)^{\top}\sum_t\boldsymbol{s}_{l-1}(t)+\sum_t{b}_{l,i}^{SNN}}{T\cdot v_{th}}-\frac{v_{l,i}(T)}{T\cdot v_{th}}.
\end{equation}
Note that the last term of (\ref{vt_fr}) satisfies that
$\frac{v_{l,i}(T)}{T\cdot v_{th}}\ll 1$
since ${v_{l,i}(T)}<v_{th}$. Hence, it can be 
ignored for sufficiently large time-steps as
\begin{equation}
\label{vt_fr2}
r_{l,i}(T)\approx\frac{\left(\boldsymbol{W}_{l-1,i}^{SNN}\right)^{\top}\sum_t\boldsymbol{s}_{l-1}(t)+\sum_t{b}_{l,i}^{SNN}}{T\cdot v_{th}}.
\end{equation}

Then, by comparing equation (\ref{relu2}) and equation (\ref{vt_fr2}), 
and with $0\leq r_{l,i}(T)\leq1$,
we can get the mapping between ANN and SNN to
convert the weights of an ANN with ReLU activation 
to an SNN with IF neurons \cite{ann2snn1,ann2snn2}. 

\subsection{Adaptive Activation Scaling for Online Adaptation}\label{AAS}

Given a pre-trained source model, 
e.g., an SNN model from ANN-SNN conversion as described in the last subsection,
our goal is to update the model on test data to improve its performance on the test data.
This can be straightforwardly achieved by using an unsupervised 
loss and a surrogate gradient based algorithm,
e.g., the BPTT algorithm introduced in Section II-A or its variants.
However, intensive experiments show that, 
directly updating the SNN model in this way can only 
achieve a performance far inferior to ANN adaption methods on cross-domain data.

\begin{figure*}[!t]%
    \centering
    \subfloat[Firing rate distribution on source data before model adaptation]{
        \includegraphics[width=0.45\linewidth]{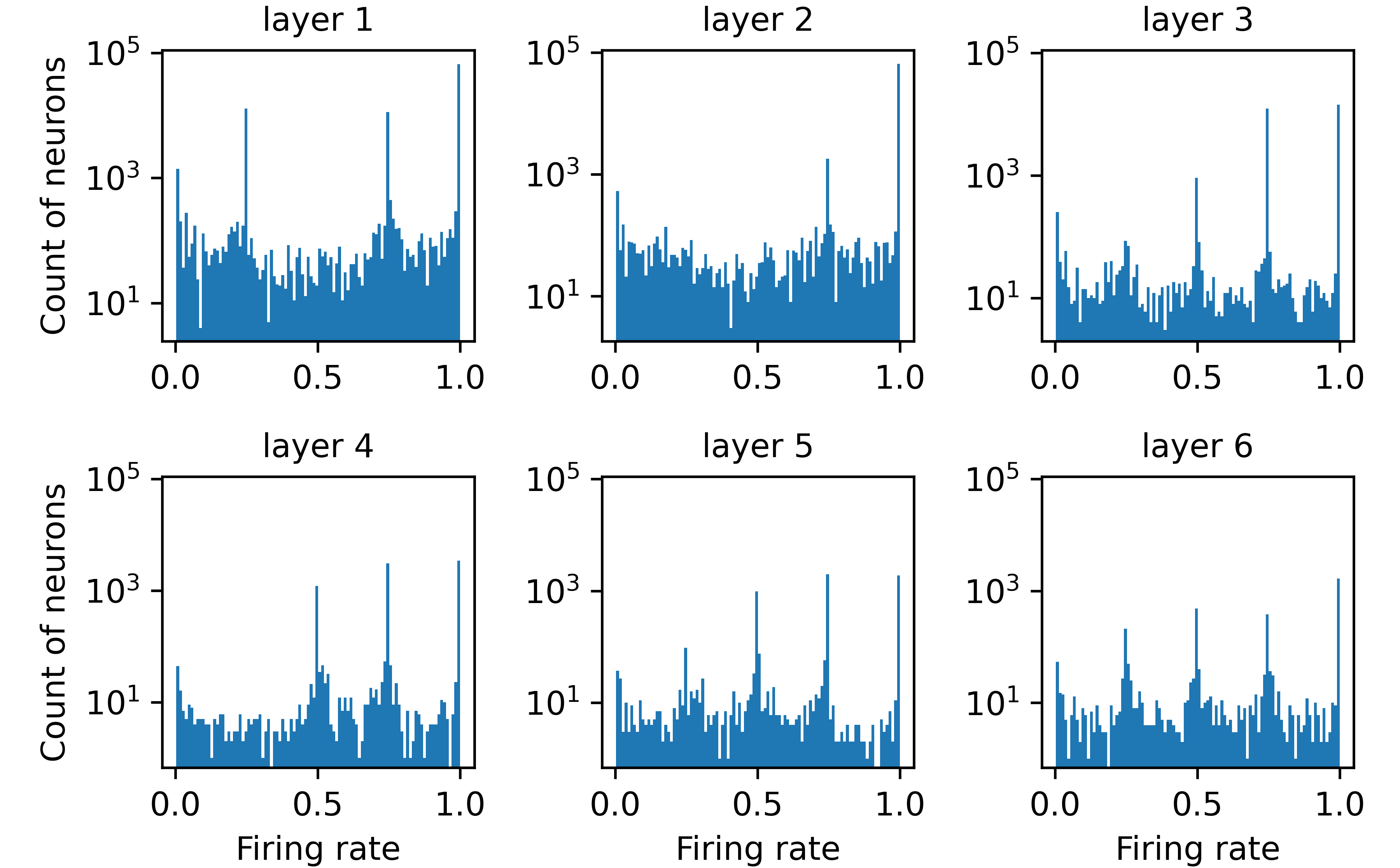}
        }~~
    \subfloat[Firing rate distribution on cross-domain test data before model adaptation]{
        \includegraphics[width=0.45\linewidth]{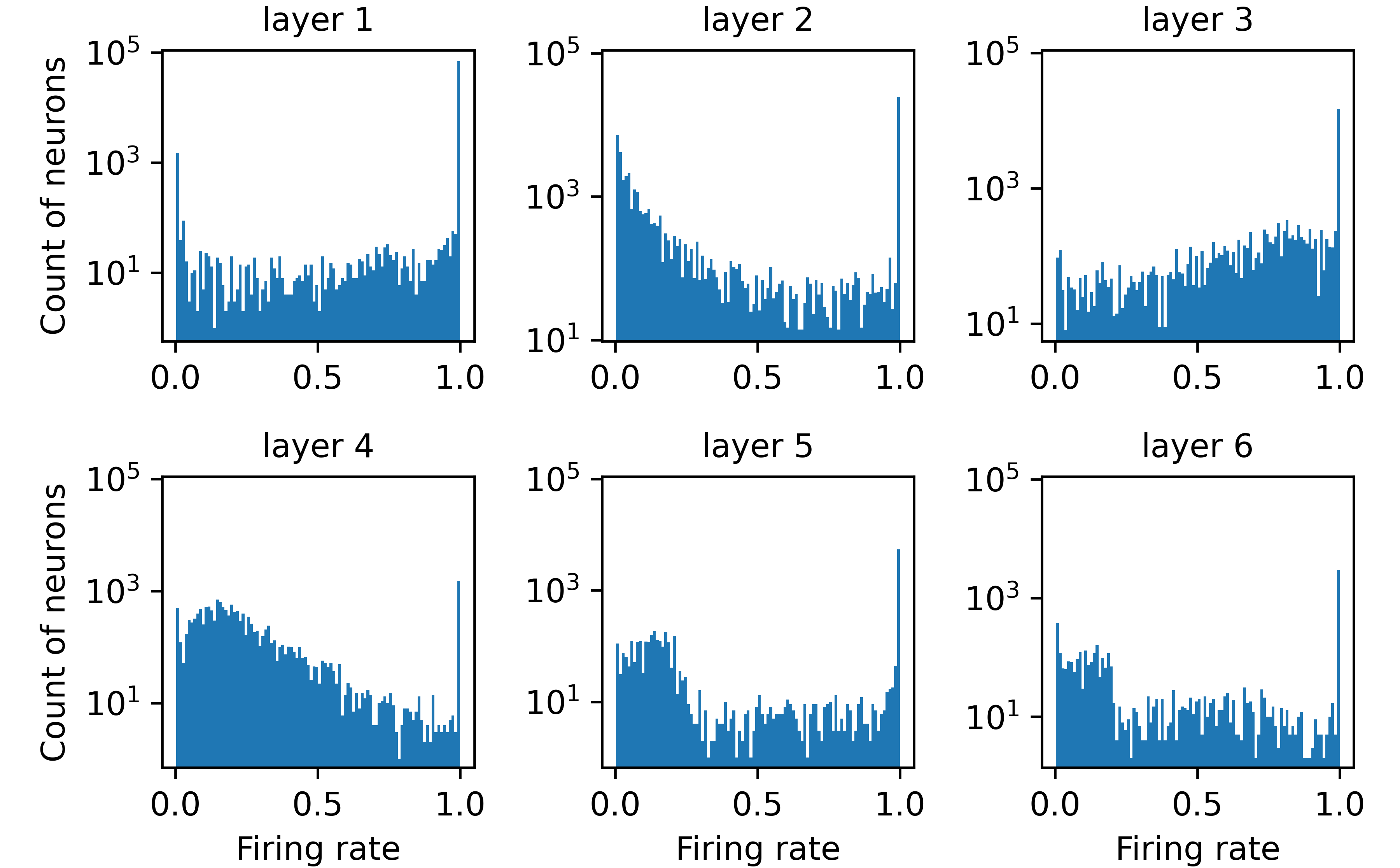}
        }\\
    \subfloat[Firing rate distribution on cross-domain test data after model adaptation without using adaptive activation scaling]{
        \includegraphics[width=0.45\linewidth]{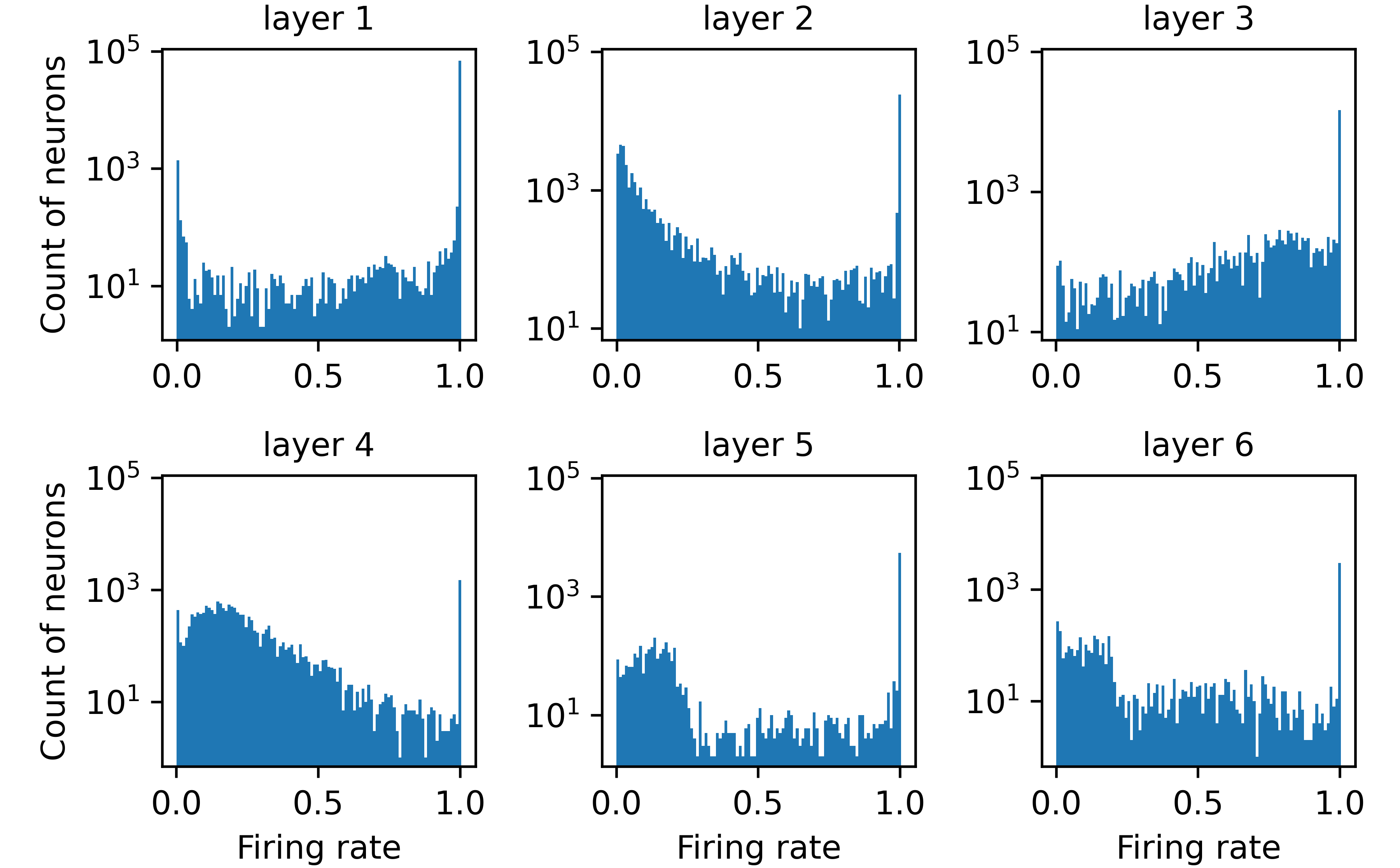}
        }~~
    \subfloat[Firing rate distribution on cross-domain test data after model adaptation using adaptive activation scaling]{
    \includegraphics[width=0.45\linewidth]{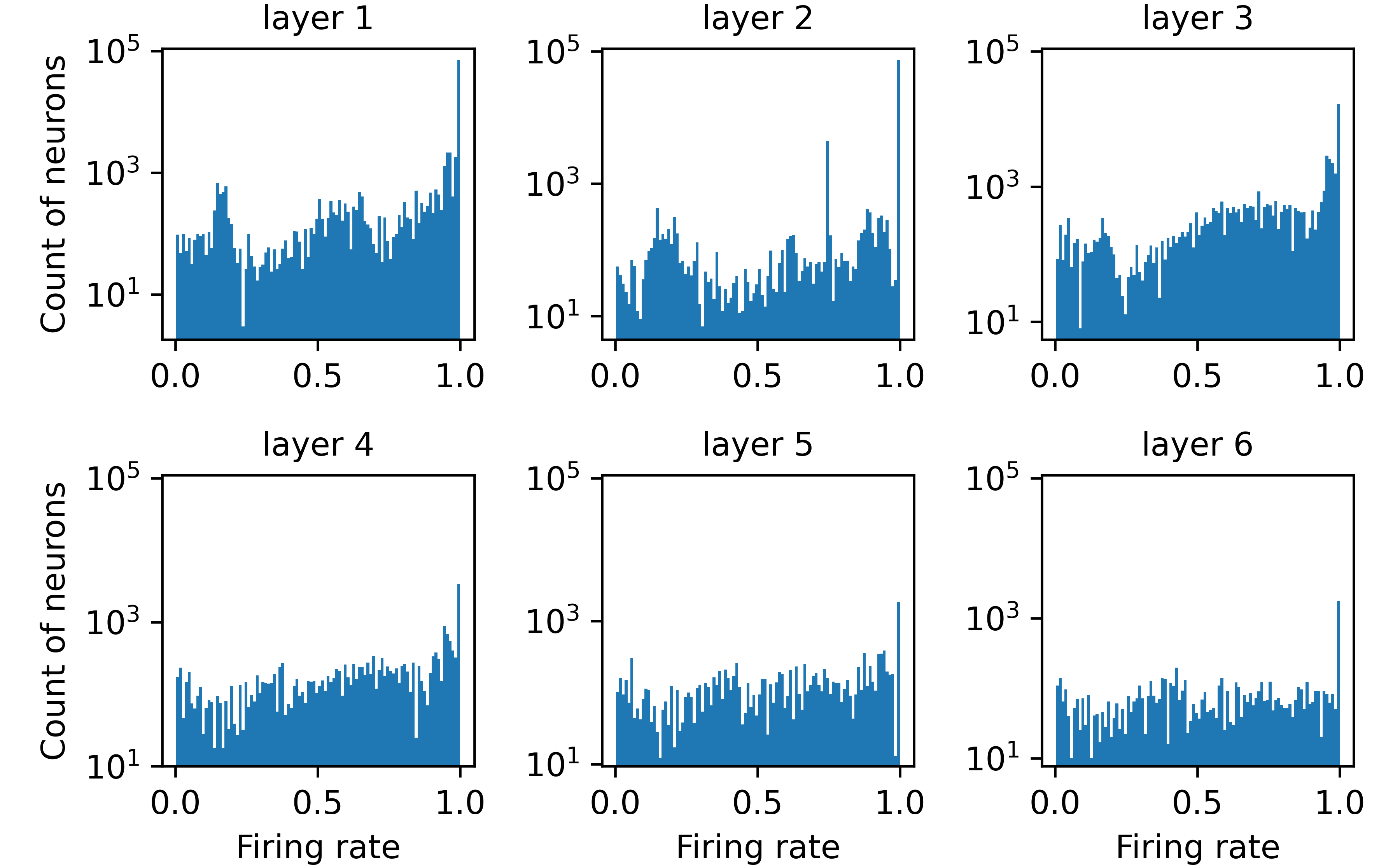}
    }
    \caption{The distribution (histograms) of firing rate of a spiking VGG16 model in cross-domain adaptation. The x-axis represents the firing rate from 0 to 1, and the y-axis (shown in log-scale) represents the count of neurons at each firing rate. 
    The firing rate distribution of the model:  {(a)} on source clean data before model adaptation, {(b)} on cross-domain test data before model adaptation, {(c)} on cross-domain test data after model adaptation without using adaptive activation scaling, {(d)} on cross-domain test data after model adaptation using adaptive activation scaling.
    }
\label{clip_histfig}
\end{figure*}

To bridge the large performance gap between 
the SNN and ANN adaptation on cross-domain data, we propose an 
adaptive activation scaling scheme for SNN adaptation to substantially
enhance the performance of SNN adaptation. 
This scheme is designed based on an analysis of the firing rate distribution 
of a spiking network in adapting on cross-domain data.
Figs. \ref{clip_histfig} shows the distribution (histograms) of the firing rate 
of different layers in a spiking VGG16 model in cross-domain adaptation. 
{The source domain dataset is RSSCN7,  
whilst the test data is simulated under cloudy weather condition}.
Detailed setting is provided in Section \ref{Experimental_Setup}. 
From Figs. \ref{clip_histfig}(a) and \ref{clip_histfig}(b), it can be seen that
the firing rate distribution of the source model has a shift on 
the corrupted test data compared to that on the source clean data.
This shift in firing rate distribution indicates that,
in the presence of domain shift of the test data,
the overall firing rate tends to drop and concentrate towards lower values.
Similar phenomenon can be observed after the model adaptation on the test data, 
as shown in Figs. \ref{clip_histfig}(c).
The shift of firing rate distribution towards lower values is detrimental to
the model's performance. Specifically, it leads to a non-uniformity 
of the firing rate distribution. For a fixed number $T$ of time-steps, 
this non-uniformity would significantly degrade the quantization accuracy
of spiking representation, and consequently affects the performance of the model.
This would be more prominent for a small number of time-steps. 

To address this problem, we propose an 
adaptive activation scaling scheme to adaptively adjust the
distribution of firing rate in the adaptation process.
Specifically, we consider an additional activation clip parameter for each layer
and dynamically adjust it to control the 
firing rate distribution. The clip parameter 
is learned in the adaptation process, 
which is expected to appropriately
compress the range of neuron activation and 
thus mitigate the non-uniformity problem of firing rate.

For a clip parameter $\alpha_l$ for layer $l$,
the neuron activation $a_l$ of layer $l$ is clipped in the feed-forward computation as
\begin{equation}
\label{aas}
\overline{a}_l = \mathrm{max}(a_l, \alpha_l) =
\begin{cases}\alpha_l, & \mathrm{if~} a_l \geq \alpha_l, \\
a_l, & \mathrm{else}.
\end{cases}
\end{equation}
Let $\boldsymbol{\alpha}=[\alpha_1,\cdots,\alpha_L]^\top$ collects the clip 
parameters of all the $L$ layers. 
As the goal is to appropriately compress the range of neuron activation,
we consider a joint loss for optimizing the clip parameters as
\begin{equation}
\label{clip_up}
\min_{\boldsymbol{\alpha}} \mathcal{L(\boldsymbol{\omega},\boldsymbol{\alpha})}+\epsilon \|\boldsymbol{\alpha}\|^2,
\end{equation}
where $\boldsymbol{\omega}$ collects all the parameters of the network other than $\boldsymbol{\alpha}$, $\mathcal{L}$ is the loss function, 
$\epsilon>0$ is an $\ell_2$-regularization coefﬁcient.
In the formulation (\ref{clip_up}), the $\ell_2$-regularization
is used to decrease the values of $\boldsymbol{\alpha}$. 
While an excessive decrease of the clip parameters may
lead to performance deterioration of the model, the first term $\mathcal{L(\boldsymbol{\omega},\boldsymbol{\alpha})}$ can help to
prevent the performance deterioration.
With the formulation (\ref{clip_up}), 
in the adaptation process, the clip parameters can be updated together 
with other model parameters.
For example, using SGD, the clip parameter $\alpha_l$ of layer $l$
can be updated as
\begin{equation}
\label{clip_optim}
\alpha_l(k+1)=\alpha_l(k)-\eta\left(\epsilon\alpha_l-\eta\frac{\partial \mathcal{L}}{\partial\alpha_l}\right),%=\alpha-\eta\epsilon\alpha-\eta\frac{\partial{a_{aas}}}{\partial\alpha}\frac{\partial L}{\partial{a_{aas}}}
\end{equation}
where $\eta$ is a learning rate, 

Since the firing rate of a spiking neuron can be 
represented as its activation value divided by the 
maximum activation of its residing layer,
the decrease of $\alpha_l$ is equivalent to the increase of 
the firing rate of the neurons in the layer.
Consequently, using this adaptive activation scaling scheme,
we can adjust the range of neuron activation and 
mitigate the non-uniformity problem of firing rate 
in the case of data distribution shift.
This is verified in Fig. \ref{clip_histfig}(d) that,
using the proposed adaptive activation scaling scheme,
the firing rate distribution of a spiking model after
adaptation on cross-domain test data becomes more uniform.
This would be more prominent for a small number of time-steps.

The proposed mechanism endows an SNN the ability to adjust the firing rate during adaptation,
which is crucial for quickly adapting to changing data distributions.
As will be shown later in experiments,
this mechanism substantially improve the performance of SNN adaptation,
especially at low time-steps. 

%介绍带温度系数的无监督最小化熵损失
\subsection{Unsupervised Online Adaptation Learning for SNN}{\label{online-alg}}

Consider the applications of real-world on-device adaptation for remote sensing,
an adaptation method is expected to be unsupervised without 
the requirement of labeled data. Meanwhile, the method should be 
highly-efficient, allowing operation on edge devices with limited computing and memory resources. This section presents an adaptation method to fulfill 
these two desiderata.

To achieve source-free unsupervised domain adaptation,
fully test time domain adaptation (TTA) has recently received 
much research attention \cite{liang2020we, liang2024comprehensive, zhao2023pitfallstesttimeadaptation,sun2020test, liu2021ttt++,wang2021tent, niu2022efficient, hong2023mecta}, which enables a pre-trained model to adapt to cross-domain test data in an unsupervised manner.
This paper follows the fully TTA works as they are more suitable for
on-device processing.
Typically, given a pre-trained source model, 
the model is adapted on unlabeled test samples based on
unsupervised losses such as the entropy of the model prediction \cite{wang2021tent}.

Following these works, we also use the prediction entropy of the model as 
the objective. 
Consider the temporal dynamic characteristics of SNN, the network's output
is accumulated over multiple time steps. 
Specifically, for a model with $T$ time-step outputs,
let $\mathbf{v}^L[t]$ denote the membrane potential of the last  layer (the $L$-th layer) at time-step $t$,
the final model prediction is calculated as $\hat{\bf{y}}=\frac{1}{T}\sum_{t=1}^{T}\mathbf{v}^L[t]\in\mathbb{R}^C$,
where $C$ is the number of classes for a classification task. 
With the model prediction $\hat{\bf{y}}$,
%Entropy \cite{entropy}, as an unsupervised objective, 
%can effectively achieve our goals since it relies solely 
%on predictions rather than annotations. Based on that, we design an unsupervised entropy loss function for our SNN model. Consider $\hat{y}$ as the model prediction, 
the entropy loss can be written as
\begin{equation}\label{entropy}
\mathcal{L}_e(\hat{\bf{y}})=-\sum_{c} {{p(\hat{\bf{y}}) \log p(\hat{\bf{y}})}},
\end{equation}
where $p(\hat{\bf{y}})=\mathrm{softmax}(\hat{\bf{y}})\in\mathbb{R}^C$ 
stands for the normalized probability expression of the output. 
Basically, the entropy $\mathcal{L}_e(\hat{\bf{y}})$ measures the
uncertainty of the model output, of which a smaller value indicates a lower uncertainty.
%Minimize the entropy loss is to minimize the uncertainty .

%Due to the temporal characteristics of SNNs, the network's output is accumulated over multiple time steps. So the model prediction is based on firing rate: $\hat{y}=\frac{1}{T}\sum_{t=1}^{T}\mathbf{s}[t]$, where $T$ is the time steps and ${s}[t]$ is the spike at the last spiking layer. 

With the unsupervised entropy loss (\ref{entropy}),
a gradient based algorithm such BPTT can be used to update the model.
However, as mentioned in Section \ref{IFBPTT}, using the standard 
BPTT for an SNN model involves unrolling the network $T$ times along the temporal dimension
and requires the maintenance of the computational graph for all previous time steps to backpropagate errors through time.
This incurs intensive computational and memory costs that 
is prohibitive for deployment on edge-devices.

%However, as mentioned in Section \ref{IFBPTT}, the standard BPTT for an SNN model containing IF neurons involves unrolling the network into a virtual feedforward network, requiring the maintenance of the computational graph for all previous time steps to backpropagate through time. This temporal gradient dependency leads to a linear growth in memory usage to store the computational graph, which significantly diminishes the brain-like efficiency of the SNN.

Inspired by the biologically plausible "eligibility traces" mechanism,
a highly-efficient credit assignment algorithm for SNN has been proposed in
\cite{trace1}. This bio-plausible algorithm integrates the local "eligibility traces"
of neurons and global top-down signal to design an efficient credit assignment algorithm
for spiking networks that does not require unrolling of the network in time.
It only involves forward-in-time computation, and thus substantially reduces 
the complexity of BPTT, from $O(n^4)$ to $O(n^2)$ for $n$ neurons.
More recently, the effectiveness of this method has been demonstrated in deep SNNs \cite{online}. 

%without backpropagation through $\frac{\partial\mathbf{u}^{l+1}[i+1]}{\partial\mathbf{s}^{l+1}[i]}$(the red gradient path in figure (\ref{eprop})).

\begin{figure}[!t] %图 
\center{\includegraphics[width=8.5cm]{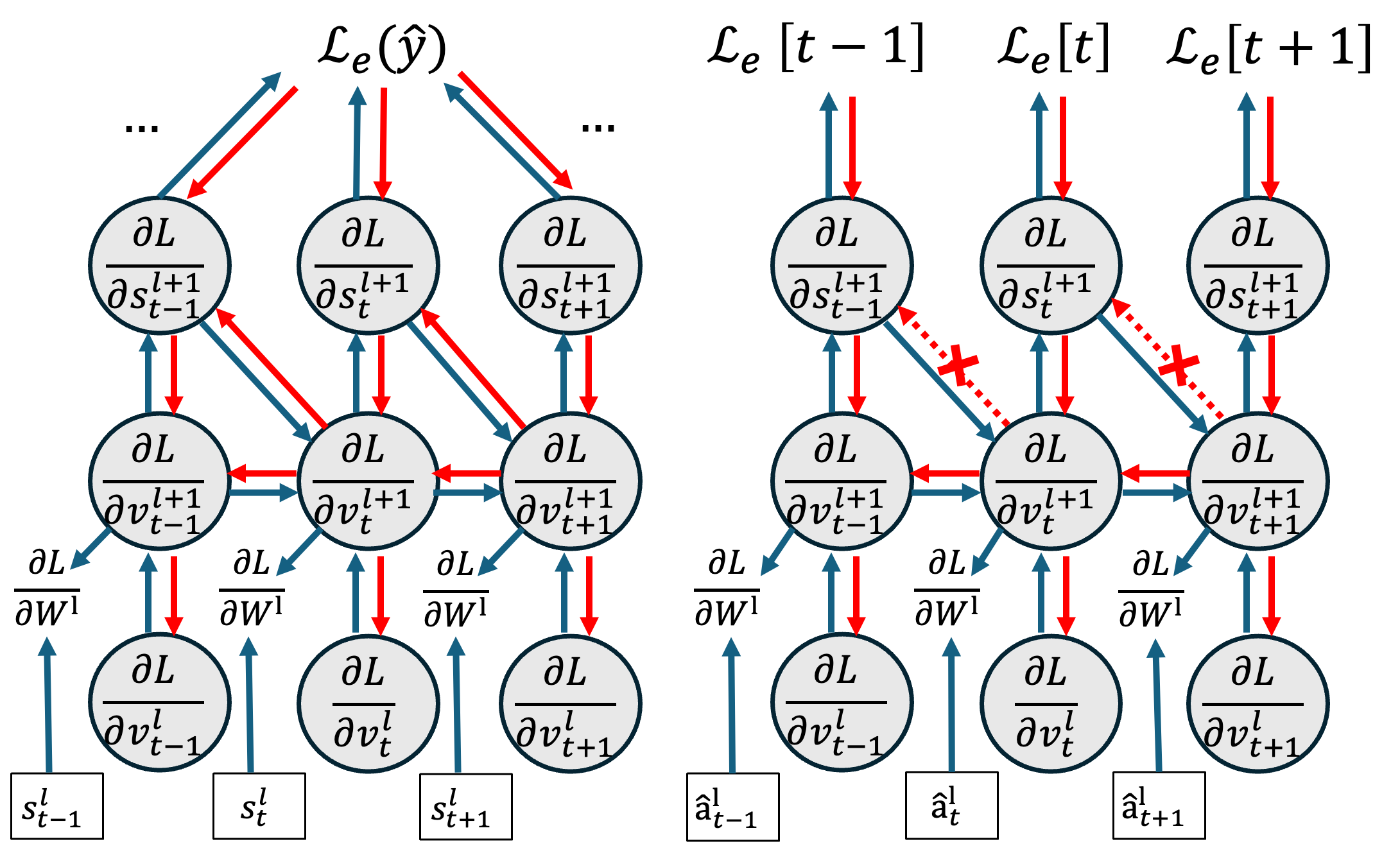}}
\caption{Illustration of the forward and backward computation of (a) BPTT and (b) the approximated online algorithm which  can compute the gradient forward-in-time.
Backward gradients are drawn in red color.
}
\label{eprop}
\end{figure}

Specifically, as illustrated in Fig. \ref{eprop}, 
this algorithm ignores the term
$\frac{\partial\mathbf{v}^{l+1}[k+1]}{\partial\mathbf{s}^{l+1}[k]}\frac{\partial\mathbf{s}^{l+1}[k]}{\partial\mathbf{v}^{l+1}[k]}$
in the full gradient computation of BPTT (\ref{bptt}).
It is equivalent to partially decouple the computation graph 
of BPTT in the temporal domain by cutting off the gradient paths $\frac{\partial\mathbf{u}^{l+1}[t+1]}{\partial\mathbf{s}^{l+1}[t]}$,
which is corresponding to reset operation of spiking emission on the membrane potential.
After partially decoupling the temporal dependencies, an approximated gradient computation of the full BPTT (\ref{bptt}) is given by \cite{online}:

\begin{equation}
\label{ottt}
\frac{\partial \mathcal{L}}{\partial\mathbf{W}^l}
=\sum_{t=1}^T\frac{\partial \mathcal{L}[t]}{\partial\mathbf{s}^{l+1}[t]}\frac{\partial\mathbf{s}^{l+1}[t]}{\partial\mathbf{v}^{l+1}[t]}\sum_{t^\prime\leq t}\frac{\partial\mathbf{v}^{l+1}[t^\prime]}{\partial\mathbf{W}^l}.
\end{equation}
This simplified algorithm can be implemented without unfolding the network.
Specifically, in implementation, it only needs to track the presynaptic activities
for each neuron in forward-in-time computation as $\hat{\mathbf{a}}^l[t\!+\!1]\! =\! \hat{\mathbf{a}}^l[t]+\mathbf{s}^l[t\!+\!1]= \sum_{t^\prime\leq t}\mathbf{s}^l[t^\prime]$.
Then, denote $\mathbf{g}_{\mathbf{v}^{l+1}}[t]\!=\!\frac{\partial \mathcal{L}(t)}{\partial\mathbf{s}^{l+1}[t]}\frac{\partial\mathbf{s}^{l+1}[t]}{\partial\mathbf{v}^{l+1}[t]}$ which dose not contain temporal dependencies, the gradient of $\mathcal{L}(t)$ with respect to $\mathbf{W}^l$, denoted by $\nabla_{\mathbf{W}^l}\mathcal{L}[t]$, can be computed at each time step independently as $\nabla_{\mathbf{W}^l}\mathcal{L}[t]=\left(\mathbf{g}_{\mathbf{v}^{l+1}}[t]\right)^\top\hat{\mathbf{a}}^l[t]^\top $.
Therefore, the gradient can be computed in a forward-in-time manner without unrolling the network, 
which substantially reduces the computation and memory load. In comparison, for a time-step $T$, the BPTT algorithm requires unrolling of the network $T$ times, of which the memory cost increases linearly with the time-step number $T$.

In this algorithm, instantaneous loss $\mathcal{L}(t)$ is used.
To apply the unsupervised entropy loss to this online algorithm,
we define instantaneous entropy loss for each time-step as
\begin{equation} 
\label{instloss}
\mathcal{L}_e[t]=-\sum_cp\left({{\bf{v}}^L[t]}\right)\log p\left({{\bf{v}}^L[t]}\right),
\end{equation}
where $p({{\bf{v}}^L[t]})=\mathrm{softmax}({{\bf{v}}^L[t]})\in\mathbb{R}^C$, $\mathbf{v}^L[t]$ denotes the membrane potential of the last layer (the $L$-th layer) at time-step $t$.

In this way, instantaneous entropy loss is defined at each
time-step, which does not require accumulation across time steps. It is used to generates an unsupervised error signal at each time step, enabling the SNN model to perform unsupervised online adaptation learning.
As shown in Fig. \ref{eprop}, this algorithm in fact approximates BPTT with gradients be calculated instantaneously at each time step to update the weights. By this means, it does not require maintaining the unfolded computational graph in the temporal dimension and only requires constant memory cost agnostic to time steps. 

%温度系数
Moreover, extensive experimental studies show that, when using an SNN model converted from an ANN model that is trained by backpropagation rather than the approximated online training algorithm, using a proper temperature parameter in calculating the model prediction benefits the entropy minimization based model adaptation. 
With a temperature scaling, the model prediction can be expressed as   
\begin{equation}\label{temperature}
%p(s[t])=\mathrm{softmax}\left(\frac{{s}(t)}{{\tau}}\right),
p({{\bf{v}}^L[t]})=\mathrm{softmax}\left(\frac{{{\bf{v}}^L[t]}}{{\tau}}\right),
\end{equation}
where $\tau$ is a temperature coefficient, which smooths the prediction probability. %At every iteration, the smoothed distribution is used for loss calculation.

\subsection{Confidence Based Instance Weighting for the Detection Task}{\label{detectionloss}}

To adapt a detection model on unlabeled test data is more challenging than that in the classification and segmentation tasks.
Intensive experiment studies show that, directly applying 
the entropy minimization based method to adapt a detection
model can even yield worse performance than the feed-forward only BN method, as will be shown later in TABLE \ref{tab:detection}.
In this section we propose a confidence based instance weighting 
scheme to significantly improve the adaptation performance on 
the detection task.

As this work considers on-device perception processing for remote sensing, 
we mainly focus on one-stage detection models, e.g. YOLO \cite{rsyolo1,rsyolo2},
which are more efficient than two-stage models and hence more suitable for on-device deployment.
Unlike the classification and semantic segmentation tasks, the detection task poses unique challenges as it involves additional bounding box regression and filtering of redundant predicted boxes. Mainstream detection models typically generate a large number of bounding box predictions, many of which may be low-quality or false positives, necessitating post-processing procedures like non-maximum suppression to eliminate low-quality or redundant boxes. Accordingly, in frameworks like YOLO, a large number of predictions are often filtered out. Therefore, for adapting an one-stage detection model, directly applying classification entropy minimization to all the predicted instances without selection cannot achieve satisfactory performance and may even be detrimental to the performance. %may not effectively optimize the model for high performance in this context. %To address this, we have introduced a novel combined unsupervised entropy loss to guide the detection model.
 
The rationale behind the proposed method is explained as follows.
Typically, a detection model generates a large number of predicted instances,
in which a large portion are noisy and unreliable, 
we design a confidence-based weighting scheme that 
filters out low-confidence ones in the predicted instances, 
and selectively use high-confidence instances to compute entropy loss 
for adaptation. %The confidence of a predicted instance is .
%Previous studies have shown that networks can effectively learn from noisy labels through negative learning. Drawing inspiration from this, we designed a confidence-based positive/negative combination entropy loss. This approach allows the model to simultaneously learn from both high-confidence reliable samples and low-confidence samples. This enables the model to increasingly differentiate between noisy samples and high-confidence samples, thereby enhancing the detection model's adaptive performance. Importantly, our method is versatile and designed not only for YOLO but also compatible with other similar detection frameworks.
Specifically, we design a weighting function based on the confidence scores
produced by the detection head of a one-stage model to assign weights for the
predicted instances in calculating the entropy loss. 
For a test image sample, 
denote ${\mathbf{P}}_{cf}\in\mathbb{R}^{H\times W}$ as the confidence score
output by the detection head, %and ${\mathbf{P}}_{cl}\in\mathbb{R}^{H\times W \times C}$ as the corresponding multi-class prediction, 
where $H$ and $W$ denote the feature map size of the last layer 
of the detection head. 
${\mathbf{P}}_{cf}$ contains the confidence score
of the presence of an object at each feature in the $H\times W$ feature map,
and ${\mathbf{P}}_{cl}$ contains the corresponding class prediction on $C$ classes at each feature.
Then, based on the confidence scores,
a weighting function $\boldsymbol{\omega}$ is calculated as
\begin{equation}
\label{weight}
\boldsymbol{\omega}(\mathbf{P}_{cf})=\frac1{1+e^{\delta(\mathbf{P}_{cf}-\tau_1)}}+\frac1{1+e^{-\delta(\mathbf{P}_{cf}-\tau_2)}},
\end{equation}
where $\tau_1$ and $\tau_2$ specify the boundaries that 
distinguishes between high confidence and low confidence.
The parameter $\delta$ controls the steepness of the weighting function.
In this weighting function, given a confidence score ${\mathbf{P}}_{cf}(i,j)$ 
of the $(i,j)$-th feature of the last feature map, 
it is considered to be of low confidence if $\tau_1<{\mathbf{P}}_{cf}(i,j)<\tau_2$,
otherwise it is considered to be of high confidence.
Intuitively, ${\mathbf{P}}_{cf}(i,j)<\tau_1$ indicates a high confidence of no object,
while ${\mathbf{P}}_{cf}(i,j)>\tau_2$ indicates a high confidence of the presence of an object.
Therefore, formula (\ref{weight}) assigns low weight for $\tau_1<{\mathbf{P}}_{cf}(i,j)<\tau_2$ to filter out low-confidence instances. 

With the weighting function (\ref{weight}), 
we define an instantaneous entropy loss for online 
optimizing the detection model as %using the algorithm introduced
\begin{equation}
\label{detectloss}
\mathcal{L}_e[t]=-\sum_{i=1}^H\sum_{j=1}^W\sum_{c=1}^C 
\boldsymbol{\omega}(\mathbf{P}_{cf}(i,j))\cdot {\mathbf{P}}_{cl}(i,j,c) \log {\mathbf{P}}_{cl}(i,j,c),
\end{equation}
% \begin{equation}
% \label{detectloss}
% L(t)=-\frac1T\sum_{t=1}^TW({p}_{conf}(t)){p}_{class}(t)\log {p}_{class}(t)
% \end{equation}
where ${\mathbf{P}}_{cl}\in\mathbb{R}^{H\times W \times C}$ contains the multi-class predictions by the detection head at each feature of the last feature map. 
This approach optimizes the model only based on high-confidence instances, 
which can mitigate influence of low-confidence instances.
As will be shown in experiments,
this approach can substantially improve the performance of adapting a detection model.

%介绍fully oneline adaptation任务的setting
\subsection{Framework of the Proposed Online Adaptation Method}
\label{configure}

The proposed online adaptation method aims to updates a 
pre-trained model on-the-fly in the on-device testing phase. 
This subsection provides a holistic overview of it.
Given a source SNN model for remote sensing image classification, 
semantic segmentation or detection, e.g. from ANN-SNN conversion,
we update the model parameters on test data in a online streaming manner. 
More specifically, upon receiving a batch of input test data,
the model produces predictions on this batch and, at meantime,
updates its parameters based on the unsupervised instantaneous entropy 
losses (\ref{instloss}) and (\ref{detectloss}).
In this manner, the pipeline updates the model on-the-fly. 

In the online adaptation phase, only a small portion of the model parameters
are updated. In the experiments, we only update the normalization layers,
which is sufficient for achieving satisfactory performance in adapting to
corruptions arise in the considered cross-domain remote sensing scenarios.
Besides, in the online adaptation phase, the adaptive activation scaling
scheme proposed in Section \ref{AAS} is adopted to adaptively adjust the
firing rate distribution, which introduces additional clip parameters 
to be learned in the adaptation phase.
Overall, the parameters to be updated online include the parameters in normalization 
layers and the clip parameters. For a BN normalization layer,
it typically first centers and standardizes the input $x$ as $\bar{x}=(x-\mu)/\sigma$, where $\mu$ and $\sigma$ denote the mean and standard deviation, respectively.
Then it uses an affine transformation $x^{\prime}=\gamma\bar{x}+\beta$
to transform $\Bar{x}$ with affine parameters $\gamma$ and $\beta$ for scale and shift, respectively.
In the online adaptation phase, only the affine parameters $\{\gamma,\beta\}$ of the normalization layers and the clip parameters $\{\alpha_l\}$ are updated, which account for only a very small fraction of the model. This benefits the adaptation efficiency.

The proposed framework is compatible with most deep learning models, 
such as VGG, ResNet, WideResNet, FCN, and other mainstream models.
Moreover, it is applicable to various tasks, such as the classification, 
semantic segmentation and detection tasks as considered in the experiments of this work.

\section{Experimental Setup}\label{Experimental_Setup}

This section introduces the datasets, the synthesis method to generate corrupted data to simulate real-world domain shifts in remote sensing, and the details of  experiment settings.

\subsection{Datasets}
To comprehensively evaluate the performance of the proposed method, we consider three remote sensing image processing tasks, including classification, semantic segmentation and detection tasks. For the classification task, we use the RSSCN7 \cite{rsscn7}, WHU-RS19\cite{whurs19}, and AID\cite{aid} datasets. For the semantic segmentation task, we use the WHDLD \cite{dlrsd_whdld}, \cite{whdld2} and DLRSD\cite{dlrsd_whdld}, \cite{dlrsd2} datasets. For the detection task, we use the RSOD and LEVIR datasets. Detailed information about these seven datasets are provided in Table \ref{tab:dataset}.

\begin{table*}[!t]
\centering
\caption{Details of the seven datasets used in experiments for the three tasks.}
\label{tab:dataset}
%\resizebox{\textwidth}{!}
{%
\begin{tabular}{@{}ccccccc@{}}
\toprule
Task & Dataset & Classes & Resolution & Source & Size & Number of images/targets \\ \midrule
\multirow{3}{*}{Classification} & RSSCN7 & 7 & - & Google Earth & 400*400 & 2800 \\
 & WHU-RS19 & 19 & 0.5 & Google Earth & 600*600 & 1005 \\
 & AID & 30 & 0.5-8 & Google Earth & 600*600 & 10000 \\ \midrule
\multirow{2}{*}{\begin{tabular}[c]{@{}c@{}}Semantic\\ segmentation\end{tabular}} & WHDLD & 6 & 2.0 & UCMerced & 256*256 & 4940 \\
 & DLRSD & 17 & 2.0 & USGS National Map & 256*256 & 2100 \\ \midrule
\multirow{2}{*}{Detection} & RSOD & 4 & 0.3-3.0 & Google Earth & 1044*915 & 6950 \\
 & LEVIR & 3 & 0.2-1.0 & Google Earth & 800*600 & 11028 \\ \bottomrule
\end{tabular}%
}
\end{table*}

\begin{figure*}[!t]%
    \centering
    \subfloat[RSSCN7]{
        \includegraphics[width=0.31\linewidth]{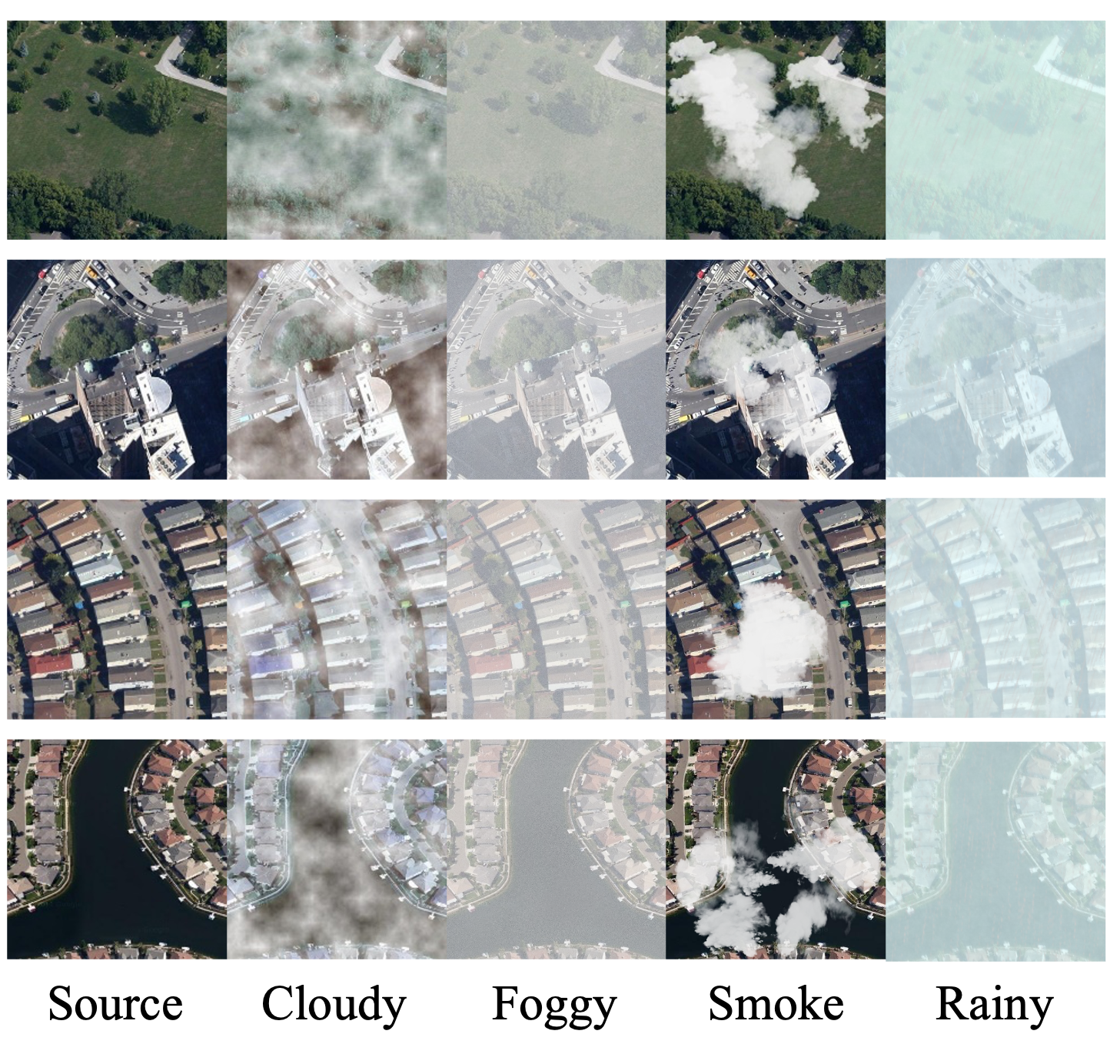}
        }%\hspace{2mm}~
    \subfloat[WHU-RS19]{
        \includegraphics[width=0.31\linewidth]{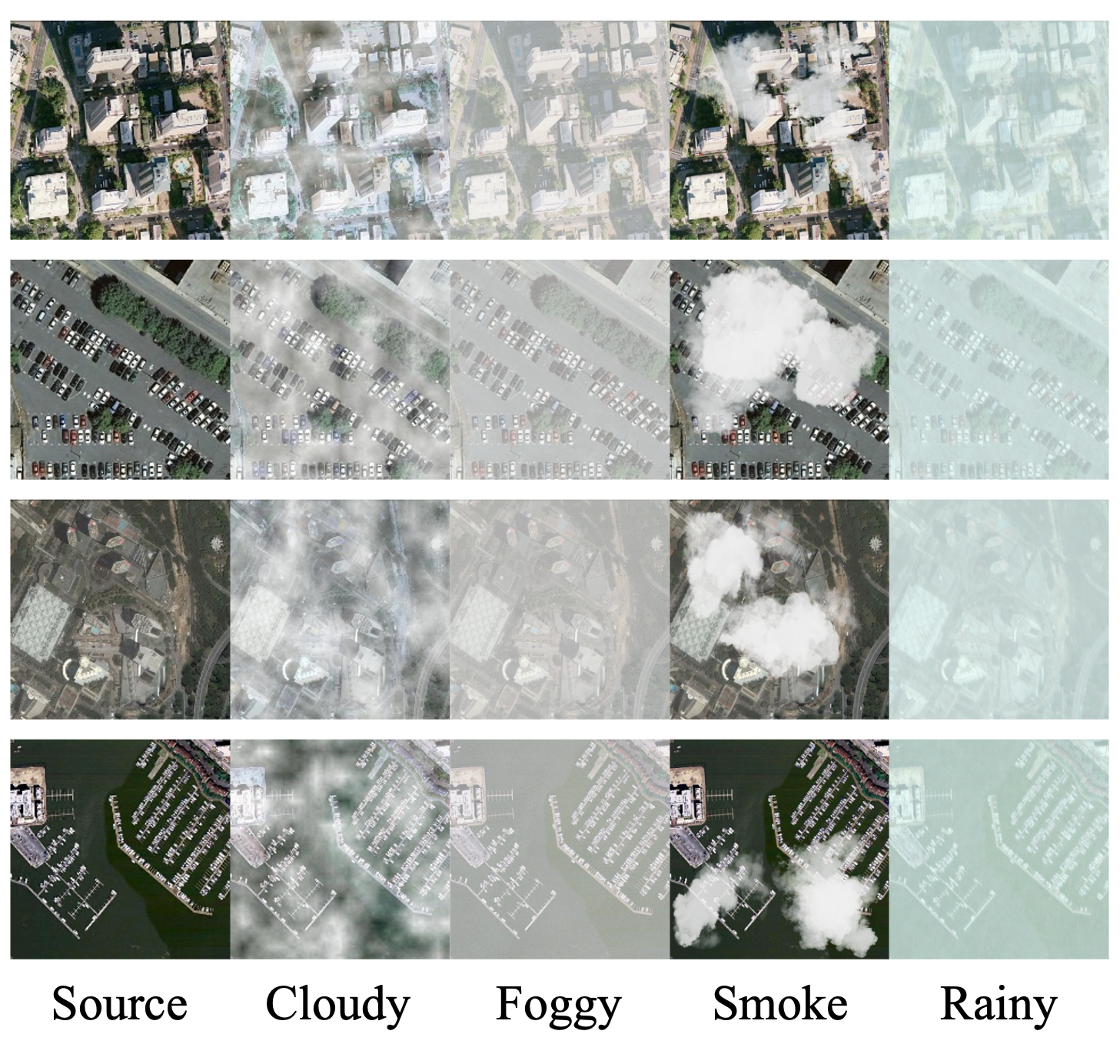}
        }%\hspace{2mm}~
    \subfloat[AID]{
        \includegraphics[width=0.31\linewidth]{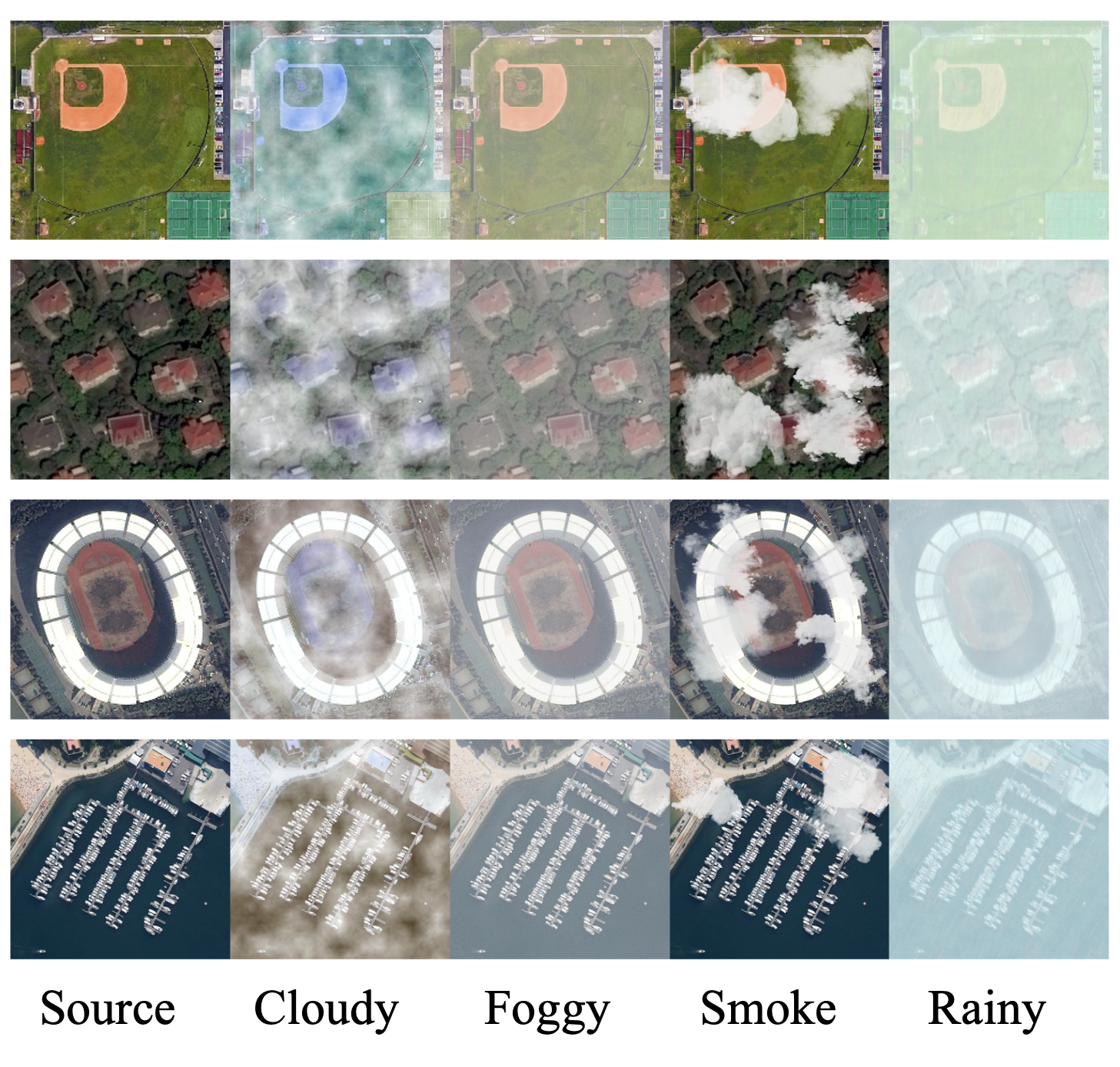}
        }%\hspace{2mm}~
    \caption{Examples of source clean images and their  corrupted versions under four types of weather conditions for each dataset. (a) The RSSCN7 dataset. (b) The WHU-RS19 dataset. (c) The AID dataset
    }
    \label{weather}
\end{figure*}

\subsection{Method for Synthesizing Corrupted Data in Remote Sensing Scenarios}

Due to the difficulty in collecting real-word datasets under different environmental variations and weather changes, and the lack of open-source datasets, we use widely accepted synthesis methods in the field of remote sensing image processing to generate corrupted data for experimental evaluation. 
To realistically simulate real-world data domain shift scenarios in remote sensing, we consider four typical weather conditions in synthesizing data corruption, which are cloudy, foggy, smoke, and rainy weather conditions.

For cloudy condition, we generate synthetic cloudy image $I_c$ by summing multi-scale random noise images \cite{cloud}. The  synthesis process can be expressed as 
\begin{equation}I_c=\sum_s\Psi\left(Rand(2^s)\right)/2^s,
\end{equation}
where $Rand(2^s)$ denotes a randomizing function that produces random noise with an image size of $2^s$, and $\Psi$ denotes the operator that resizes the random noise to the cloud image size. The scale factor $s$ is a natural number ranging from 1 to $log_2(N)$, where $N$ is the cloudy image size.

For the foggy condition, we use the diamond-square algorithm to render the fog\cite{foggy}. As for the smoke and rainy environments, we use the Adverse-Weather-Simulation to render images\cite{AdverseWeatherSimulation}. Example samples of source images and rendered synthetic corrupted images with different weather conditions are shown in Figs. \ref{weather}.

\subsection{Implementation Detail and Experimental Settings}

In the experiments, we consider VGG16 \cite{vgg} and ResNet34 \cite{resnet} as the backbone networks for the classification task, the classic FCN \cite{fcn} network as the backbone network for the semantic segmentation task, and classic YOLOv3 \cite{yolov3} with Darknet backbone as the detection network.  

First, we randomly split the clean source datasets into training and testing sets by 80\% and 20\% for each category, and then render the testing sets into corrupted testing sets using the above synthesis methods. 
After training the networks on the clean source training sets, we test them with the clean source testing sets to verify that the networks perform well on the source data. 
We then convert these ANN models into SNN as introduced in Section \ref{ANN-SNN}, and configure them to update the forward and backward propagation parameters of specific layers, as described in Section \ref{configure}.  This process does not involve additional training or learning.%, it can be completed immediately.

During adaptation, we feed the synthesized corrupted testing data to the SNN model. Since our method does not require source domain data, it is source-free. Meanwhile, it  updates the affine parameters of the normalization layers and the adaptive activation scaling parameters based on unsupervised entropy loss. Additionally, as the model processes on and adapts to each batch of test data in a streaming manner, updating parameters immediately after each batch input, our method is fully online and operates on-the-fly.

The experiments are implemented using the Pytorch library. 
All experiments are conducted on an NVIDIA GeForce RTX4090 GPU with 24G graphics memory.

\section{Results and Analysis}

\begin{table*}[!t]
\centering
\caption{Results of the compared adaptation methods in the classification task on the RSSCN7 dataset. Top-1 accuracy (\%) is reported.}
\label{tablersscn7}
\resizebox{\textwidth}{!}
{
\begin{tabular}{@{}ccccccccccc@{}}

\toprule
Model & Method & Model type & Source & Target & Loss & Cloudy & Foggy & Smoke & Rainy & Mean Acc\\ \midrule
\multirow{7}{*}{VGG16} & Oracle & ANN & \ding{55} & $x^t$ & - & - & - & - & - & 96.54 \\
 & Source-only & ANN & \ding{55} & $x^t$ & - & 27.32 & 43.39 & 30.89 & 36.25 & 34.46 \\ \cmidrule(l){2-11} 
 & TTT\cite{ttt} & ANN & $x^s, y^s$ & $x^t$ & $L(x^s,y^s)+L(x^s)$ & 52.56 & 63.52 & 54.58 & 58.45 & 57.28 \\
 & ADA\cite{ada} & ANN & $x^s, y^s$ & $x^t$ & $L(x^s,y^s)+L(x^s,x^t)$ & 51.56 & 61.71 & 52.18 & 57.76 & 55.80 \\
 & BN\cite{bn} & ANN & \ding{55} & $x^t$ & - & 53.54 & 68.26 & 60.58 & 65.85 & 62.06 \\
 & Tent\cite{tent} & ANN & \ding{55} & $x^t$ & $L(x^t)$ & 68.52 & 72.58 & 62.08 & 72.58 & 68.94 \\
 & Ours & SNN & \ding{55} & $x^t$ & $L(x^t)$ & \textbf{69.25} & \textbf{73.05} & \textbf{64.75} & \textbf{73.06} & \textbf{70.03} \\ \midrule
\multirow{7}{*}{ResNet34} & Oracle & ANN & \ding{55} & $x^t$ & - & - & - & - & - & 97.35 \\
 & Source-only & ANN & \ding{55} & $x^t$ & - & 37.32 & 46.25 & 36.79 & 47.68 & 42.01 \\ \cmidrule(l){2-11} 
 & TTT\cite{ttt} & ANN & $x^s, y^s$ & $x^t$ & $L(x^s,y^s)+L(x^s)$ & 53.78 & 64.58 & 55.87 & 59.54 & 58.44 \\
 & ADA\cite{ada} & ANN & $x^s, y^s$ & $x^t$ & $L(x^s,y^s)+L(x^s,x^t)$ & 52.58 & 63.47 & 55.25 & 58.58 & 57.47 \\
 & BN\cite{bn} & ANN & \ding{55} & $x^t$ & - & 62.57 & 70.46 & 61.92 & 69.52 & 66.12 \\
 & Tent\cite{tent} & ANN & \ding{55} & $x^t$ & $L(x^t)$ & 68.28 & 75.25 & 64.14 & 72.58 & 70.06 \\
 & Ours & SNN & \ding{55} & $x^t$ & $L(x^t)$ & \textbf{71.06} & \textbf{75.55} & \textbf{65.05} & \textbf{74.25} & \textbf{71.48} \\ \bottomrule
\end{tabular}
}
\end{table*}

\subsection{Compared Methods and Evaluation Metrics}

To comprehensively evaluate the proposed method in comparison with existing methods, the
following state-of-the-art unsupervised adaptation methods are compared.% The detail of these methods are shown in the following. 

\begin{enumerate}
\item{\textit{Oracle}: Oracle means a model is only trained on the labeled source domain training data while tested on source domain testing data. Since test samples are clean and independently identically distributed with the source domain training samples, the oracle model provides a upper bound of performance.}
\item{\textit{Source-only}: Source-only represents a pre-trained source model is directly tested on the corrupted data without adaptation. Its performance would drop dramatically since the model does not adapt to the test data with domain shift. It provides a lower bound of  performance and helps reflect the effectiveness of the adaptation methods.}
%\item{\textit{Source-only}: Source-only represents model is only trained on the labeled source domain data while directly tested on the corrupted data domain. The retrieval performance drops dramatically since the model cannot adapt to the changing data domain.It represents the scenario of prediction failure in the event of model collapse to provide the lower bound of baseline performance and indicate the effectiveness and reliability of other unsupervised methods.}
\item{\textit{TTT}: Test-time training (TTT)\cite{ttt}  trains a model based on joint supervised and self-supervised tasks on source domain, and then uses  the self-supervised task to train the model on test data. }
{\item{\textit{ADA}: The adversarial domain adaptation (ADA) method \cite{ada} reverses the gradients of a domain classifier on source and test data to learn domain-invariant representation.}
\item{\textit{AdaptSegNet}: AdaptSegNet \cite{adaptseg} first adopts adversarial learning in the output space for UDA semantic segmentation. Compared with adaptation at the feature level, the structured output spaces of the source and target domains contain global context and spatial similarities, which are beneficial for adaptation. In this method, a multilevel adversarial network is designed to enhance the model.}
\item{\textit{ADVENT}: Building on AdaptSegNet, ADVENT \cite{advent} calculates the entropy of the output space to avoid low-confidence predictions in the target domain. The entropy map is then fed into the discriminator for adversarial training.}}
\item{\textit{BN}: It only updates the normalization statistics \cite{bn} on test data during testing, which only involves froward computation.}
\item{\textit{Tent}: Test entropy (Tent) \cite{tent} adapts a model on test data by minimizing the entropy of the model prediction. }
\end{enumerate}
 
We exclusively evaluate TTT and ADA for the classification task, 
while AdapSegNet and ADVENT are exclusively evaluated for the semantic segmentation task. 
BN and Tent are evaluated for all the three tasks.
Compared to the classification and segmentation tasks,
the research on online adaptation for the detection task is very limited \cite{detectionTTA}.
In the experiment on the detection task, we apply the BN and Tent methods \cite{bn,tent} 
to the detection task, and compare them with the proposed method.
%{It is worth noting that no other methods besides BN and Tent have explored online adaptation for the detection task. Therefore, we compare our method only with these two approaches.}
It is also important to acknowledge that some methods deviate from strictly adhering to fully online settings, unlike our proposed method. Only BN and Tent are compared under exactly the same settings. Other methods may incorporate additional supervision or require access to source data, but our approach still show advantage compared with them.%even under these more challenging comparisons.

We use overall accuracy and intersection-over-union (IoU) to evaluate the performance of the methods on the classification task and semantic segmentation task, respectively.
% \begin{equation}\mathrm{IoU}=\frac{\mathrm{TP}}{\mathrm{TP}+\mathrm{FN}+\mathrm{FP}}\end{equation}
% where TP, FP, and FN denote the number of true positive pixels, false positive pixels, and false negative pixels, respectively. A larger metric value indicates better performance.
For the detection task, the performance of the methods are evaluated in terms of mean average precision (mAP).

\subsection{Results on the Classification Task}

\begin{table*}[!t]
\centering
\caption{Results of the compared adaptation methods in the classification task on the WHU-RS19 dataset. Top-1 accuracy (\%) is reported.}
\label{tablewhurs19}
\resizebox{\textwidth}{!}
{
\begin{tabular}{@{}ccccccccccc@{}}
\toprule
Model & Method & Model type & Source & Target & Loss & Cloudy & Foggy & Smoke & Rainy & Mean Acc\\ \midrule
\multirow{7}{*}{VGG16} & Oracle & ANN & \ding{55} & $x^t$ & - & - & - & - & - & 92.54 \\
 & Source-only & ANN & \ding{55} & $x^t$ & - & 32.45 & 42.75 & 31.80 & 47.55 & 38.64 \\ \cmidrule(l){2-11} 
 & TTT\cite{ttt} & ANN & $x^s, y^s$ & $x^t$ & $L(x^s,y^s)+L(x^s)$ & 49.65 & 55.71 & 48.32 & 57.43 & 52.78 \\
 & ADA\cite{ada} & ANN & $x^s, y^s$ & $x^t$ & $L(x^s,y^s)+L(x^s,x^t)$ & 47.25 & 53.48 & 47.57 & 56.38 & 51.17 \\
 & BN\cite{bn} & ANN & \ding{55} & $x^t$ & - & 51.35 & 54.62 & 52.83 & 63.15 & 55.49 \\
 & Tent\cite{tent} & ANN & \ding{55} & $x^t$ & $L(x^t)$ & 54.28 & 59.54 & 56.08 & 64.58 & 58.62 \\
 & Ours & SNN & \ding{55} & $x^t$ & $L(x^t)$ & \textbf{55.86} & \textbf{59.57} & \textbf{57.04} & \textbf{64.78} & \textbf{59.31} \\ \midrule
\multirow{7}{*}{ResNet34} & Oracle & ANN & \ding{55} & $x^t$ & - & - & - & - & - & 96.35 \\
 & Source-only & ANN & \ding{55} & $x^t$ & - & 33.83 & 43.60 & 32.34 & 48.26 & 39.51 \\ \cmidrule(l){2-11} 
 & TTT\cite{ttt} & ANN & $x^s, y^s$ & $x^t$ & $L(x^s,y^s)+L(x^s)$ & 51.79 & 54.48 & 49.67 & 59.67 & 53.90 \\
 & ADA\cite{ada} & ANN & $x^s, y^s$ & $x^t$ & $L(x^s,y^s)+L(x^s,x^t)$ & 48.35 & 53.85 & 48.46 & 59.58 & 52.56 \\
 & BN\cite{bn} & ANN & \ding{55} & $x^t$ & - & 52.76 & 55.73 & 53.94 & 64.25 & 56.67 \\
 & Tent\cite{tent} & ANN & \ding{55} & $x^t$ & $L(x^t)$ & 55.78 & 59.13 & 56.63 & 65.20 & 59.19 \\
 & Ours & SNN & \ding{55} & $x^t$ & $L(x^t)$ & \textbf{56.67} & \textbf{59.54} & \textbf{56.68} & \textbf{65.36} & \textbf{59.56} \\ \bottomrule
\end{tabular}
}
\end{table*}

\begin{table*}[!t]
\centering
\caption{Results of the compared adaptation methods in the classification task on the AID dataset. Top-1 accuracy (\%) is reported.}
\label{tableAID}
\resizebox{\textwidth}{!}
{
\begin{tabular}{@{}ccccccccccc@{}}
\toprule
Model & Method & Model type & Source & Target & Loss & Cloudy & Foggy & Smoke & Rainy & Mean Acc\\ \midrule
\multirow{7}{*}{VGG16} & Oracle & ANN & \ding{55} & $x^t$ & - & - & - & - & - & 91.85 \\
 & Source-only & ANN & \ding{55} & $x^t$ & - & 14.25 & 21.85 & 16.70 & 18.45 & 17.81 \\ \cmidrule(l){2-11} 
 & TTT\cite{ttt} & ANN & $x^s, y^s$ & $x^t$ & $L(x^s,y^s)+L(x^s)$ & 34.89 & 49.50 & 39.52 & 40.64 & 41.14 \\
 & ADA\cite{ada} & ANN & $x^s, y^s$ & $x^t$ & $L(x^s,y^s)+L(x^s,x^t)$ & 34.15 & 45.68 & 37.86 & 39.57 & 39.32 \\
 & BN\cite{bn} & ANN & \ding{55} & $x^t$ & - & 37.64 & 53.19 & 40.19 & 44.95 & 43.99 \\
 & Tent\cite{tent} & ANN & \ding{55} & $x^t$ & $L(x^t)$ & 42.67 & 55.68 & 41.67 & 51.58 & 47.90 \\
 & Ours & SNN & \ding{55} & $x^t$ & $L(x^t)$ & \textbf{43.37} & \textbf{55.84} & \textbf{41.99} & \textbf{52.57} & \textbf{48.44} \\ \midrule
\multirow{7}{*}{ResNet34} & Oracle & ANN & \ding{55} & $x^t$ & - & - & - & - & - & 92.15 \\
 & Source-only & ANN & \ding{55} & $x^t$ & - & 13.45 & 22.75 & 15.80 & 19.55 & 17.89 \\ \cmidrule(l){2-11} 
 & TTT\cite{ttt} & ANN & $x^s, y^s$ & $x^t$ & $L(x^s,y^s)+L(x^s)$ & 34.99 & 49.60 & 39.96 & 39.67 & 41.06 \\
 & ADA\cite{ada} & ANN & $x^s, y^s$ & $x^t$ & $L(x^s,y^s)+L(x^s,x^t)$ & 34.25 & 44.78 & 38.62 & 39.74 & 39.35 \\
 & BN\cite{bn} & ANN & \ding{55} & $x^t$ & - & 35.54 & 51.09 & 40.81 & 42.85 & 42.57 \\
 & Tent\cite{tent} & ANN & \ding{55} & $x^t$ & $L(x^t)$ & 42.57 & 56.53 & 41.89 & 52.18 & 48.29 \\
 & Ours & SNN & \ding{55} & $x^t$ & $L(x^t)$ & \textbf{43.27} & \textbf{56.74} & \textbf{42.56} & \textbf{52.47} & \textbf{48.76} \\ \bottomrule
\end{tabular}
}
\end{table*}
 
TABLE \ref{tablersscn7} presents the classification accuracy of the compared methods on the RSSCN7 dataset. For a fair comparison, all methods use the same backbone model. We consider two backbones, VGG16 and ResNet34. We report the accuracy on corrupted data in the cloudy, foggy, smoke, and rainy conditions, and present the mean overall accuracy across these four conditions. Before adaptation, both the VGG and ResNet models can achieve a high accuracy (96.54\% and 97.35\%, respectively) on the clean source data. However, both models experience a large accuracy drop of more than 50\%  on corrupted data, indicating that weather interference can lead to significant deterioration of the models' performance. It can be seen that, TTT and ADA can improve the performance by over 20\% but they additionally use source data. 

BN and Tent can raise the accuracy to over 60\%, e.g., reaching 62.06\% and 68.94\% with the VGG16 network, and 66.12\% and 70.06\% with the ResNet34 network, respectively. While BN only updates the normalization layers' statistics parameters during forward computation, Tent additionally updates the affine parameters of the normalization layers through backpropagation, which has better adaptation performance than BN. Our method using an SNN model achieves the best performance in terms of mean accuracy with both VGG16 and Resnet34 backbones, e.g., 70.03\% and 71.48\%, respectively. 

TABLE \ref{tablewhurs19} shows the results on the WHU-RS19 dataset. On the WHU-RS19 dataset, the source models without adaptation also drop over 50\% of accuracy from the oracle. The TTT method slightly outperforms ADA, while Tent slightly outperforms BN. The proposed method again achieves the best performance, e.g., 59.31\% and 59.56\% accuracy with the VGG16 and ResNet34 backbones, respectively. 
On the AID dataset, we observe a similar trend from TABLE \ref{tableAID}. Due to the increase of classes and the complexity of the image data, the source models  drop more compared to their performance on the other two datasets. While the source models achieve over 91\% accuracy on the source data, they only have an accuracy less than 20\% on corruption data, indicating that the models  fail in challenging  scenarios when facing corruption. As shown in TABLE \ref{tableAID}, with the VGG16 backbone, the TTT and ADA methods yield accuracy around 40\%, whilst BN and Tent have better performance reaching 43.99\% and 47.9\%, respectively. Our method achieves an accuracy over 48\% with both backbones.
From the results, we can observe that the overall accuracy on cloudy and smoke corruptions is lower than that on the other two corruptions. This is due to the nature that the cloudy and smoke conditions are more challenging, %which directly add extra patterns to the original images, 
in which the corruption obscures the original features and significantly increases the difficulty for adaptation. But overall, our SNN adaptation method shows strong adaptation capability and outperforms the compared adaptation methods.

\begin{table*}[!t]
\centering
\caption{Results of the compared adaptation methods in the semantic segmentation task on the DLRSD and WHDLD datasets.
Mean intersection-over-union (mIoU) (in \%) is reported.}
\label{tab:segmentation}
\resizebox{\textwidth}{!}
{%
\begin{tabular}{@{}ccccccccccc@{}}
\toprule
Dataset & Method & Model type & Source & Target & Loss & Cloudy & Foggy & Smoke & Rainy & Mean mIoU \\ \midrule
\multirow{7}{*}{DLRSD} & Oracle & ANN & \ding{55} & $x^t$ & - & - & - & - & - & 66.05 \\
 & Source-only & ANN & \ding{55} & $x^t$ & - & 17.25 & 31.75 & 15.75 & 35.75 & 25.13 \\ \cmidrule(l){2-11} 
 & AdaptSegNet\cite{adaptseg} & ANN & $x^s, y^s$ & $x^t$ & $L(x^s,y^s)+L(x^s,x^t)$ & 19.78 & 40.72 & 20.85 & 41.52 & 30.72 \\
 & ADVENT\cite{advent} & ANN & $x^s, y^s$ & $x^t$ & $L(x^s,y^s)+L(x^s,x^t)+L(x^t)$ & 21.86 & 42.18 & 23.48 & 42.85 & 32.59 \\
 & BN\cite{bn} & ANN & \ding{55} & $x^t$ & - & 27.42 & 43.79 & 28.85 & 46.72 & 36.70 \\
 & Tent\cite{tent} & ANN & \ding{55} & $x^t$ & $L(x^t)$ & 33.15 & 47.78 & 32.87 & 48.27 & 40.52 \\
 & Ours & SNN & \ding{55} & $x^t$ & $L(x^t)$ & \textbf{34.35} & \textbf{47.88} & \textbf{33.72} & \textbf{49.16} & \textbf{41.28} \\ \midrule
\multirow{7}{*}{WHDLD} & Oracle & ANN & \ding{55} & $x^t$ & - & - & - & - & - & 64.38 \\
 & Source-only & ANN & \ding{55} & $x^t$ & - & 19.17 & 29.89 & 14.06 & 31.67 & 23.70 \\ \cmidrule(l){2-11} 
 & AdaptSegNet\cite{adaptseg} & ANN & $x^s, y^s$ & $x^t$ & $L(x^s,y^s)+L(x^s,x^t)$ & 22.48 & 38.28 & 25.85 & 42.04 & 32.16 \\
 & ADVENT\cite{advent} & ANN & $x^s, y^s$ & $x^t$ & $L(x^s,y^s)+L(x^s,x^t)+L(x^t)$ & 34.82 & 40.06 & 28.95 & 43.28 & 36.78 \\
 & BN\cite{bn} & ANN & \ding{55} & $x^t$ & - & 39.74 & 45.79 & 33.87 & 51.85 & 42.81 \\
 & Tent\cite{tent} & ANN & \ding{55} & $x^t$ & $L(x^t)$ & 41.22 & 47.32 & 35.92 & 53.57 & 44.51 \\
 & Ours & SNN & \ding{55} & $x^t$ & $L(x^t)$ & \textbf{42.38} & \textbf{48.06} & \textbf{36.28} & \textbf{54.09} & \textbf{45.20} \\ \bottomrule
\end{tabular}%
}
\end{table*}

\begin{figure*}[!t]
\center{\includegraphics[width=16cm]{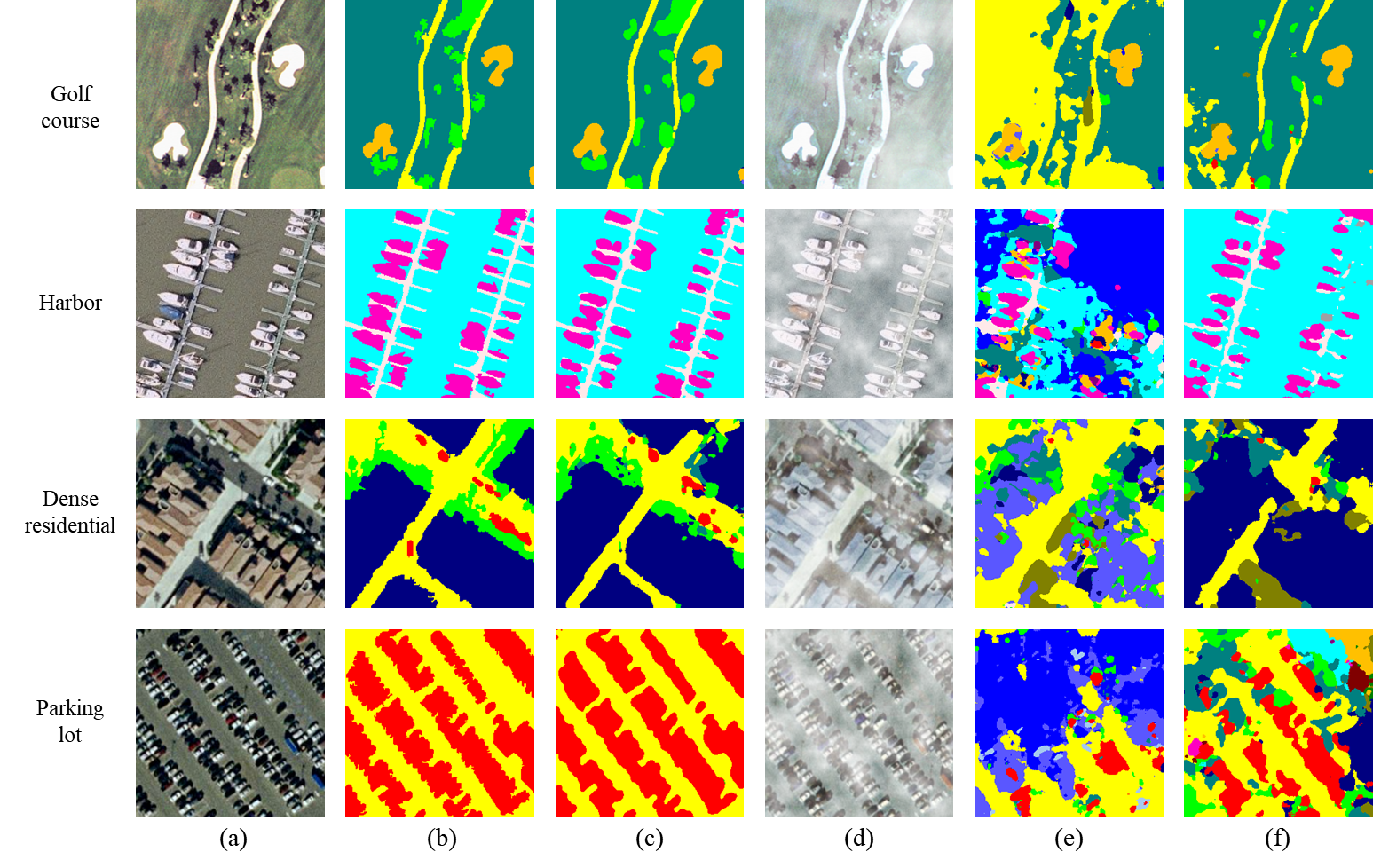}} 
\caption{Visualization of segmented results on image samples from the WHDLD dataset by the compared methods. (a) Original images. (b) Ground-truth segmentation. (c) Oracle. (d) Cloudy images. (e) Source-only model. (f) Our method.
}
\label{segmentation}
\end{figure*}

\subsection{Results on the Semantic Segmentation Task}

To demonstrate the broad applicability of the proposed method, we further conduct adaptation experiments on semantic segmentation of remote sensing images. We use FCN \cite{fcn} as the backbone network and conduct adaptation experiments on two optical remote sensing datasets, 
WHDLD \cite{dlrsd_whdld,whdld2} and DLRSD\cite{dlrsd_whdld,dlrsd2}.
TABLE \ref{tab:segmentation} shows the semantic segmentation performance of the compared methods on the WHDLD and DLRSD datasets. 

It can be seen from TABLE \ref{tab:segmentation} that, 
weather interference has a significant negative impact on the performance of segmentation models. The backbone network, which is well-trained on the source domain with over 60\% mIoU, degrades to have an mIoU barely above 20\% on corrupted test data. The degradation is especially conspicuous under the cloudy and smoke conditions. On the DLRSD dataset, the source-only method achieves only 17.25\% and 15.75\% mIoU under the cloudy and smoke conditions, respectively. Similar large degradation of the source model can also be observed on the WHDLD dataset, e.g., with an mIoU of 19.17\% and 14.06\% on the cloudy and smoke conditions, respectively.

\begin{table*}[!t]
\centering
\caption{Results of the compared adaptation methods in the detection  task on the RSOD and LEVIR datasets.
mAP (in \%) is reported.}
\label{tab:detection}
\resizebox{\textwidth}{!}
{
\begin{tabular}{@{}ccccccccccc@{}}
\toprule

Datasets & Method & Model type & Source & Target & Loss & Cloudy & Foggy & Smoke & Rainy & Mean mAP \\ \midrule
 \multirow{5}{*}{RSOD} & Oracle & ANN & \ding{55} & $x^t$ & - & - & - & - & - & 82.71 \\
 & Source-only & ANN & \ding{55} & $x^t$ & - & 12.89 & 24.92 & 11.75 & 26.80 & 19.09 \\ \cline{2-11} 
 & BN\cite{bn} & ANN & \ding{55} & $x^t$ & - & 39.45 & \textbf{45.24} & \textbf{47.34} & \textbf{48.82} & 45.21 \\
 & TENT\cite{tent} & ANN & \ding{55} & $x^t$ & $L(x^t)$ & 39.12 & 45.15 & 46.17 & 47.56 & 44.50 \\
 & Ours & SNN & \ding{55} & $x^t$ & $L(x^t)$ & \textbf{41.75} & \textbf{46.75} & \textbf{49.85} & \textbf{49.83} & \textbf{47.05} \\ \hline
\multirow{5}{*}{LEVIR} & Oracle & ANN & \ding{55} & $x^t$ & - & - & - & - & - & 77.84 \\
 & Source-only & ANN & \ding{55} & $x^t$ & - & 11.85 & 25.76 & 13.75 & 23.54 & 18.73 \\ \cline{2-11} 
 & BN\cite{bn} & ANN & \ding{55} & $x^t$ & - & 35.48 & 43.82 & 38.75 & 45.78 & 40.96 \\
 & TENT\cite{tent} & ANN & \ding{55} & $x^t$ & $L(x^t)$ & 34.05 & 42.76 & 36.07 & 44.25 & 39.28 \\
 & Ours & SNN & \ding{55} & $x^t$ & $L(x^t)$ & \textbf{37.81} & \textbf{44.25} & \textbf{39.36} & \textbf{47.06} & \textbf{42.12} \\ \hline
\end{tabular}%
}
\end{table*}

\begin{figure*}[!t]
\center{\includegraphics[width=16cm]{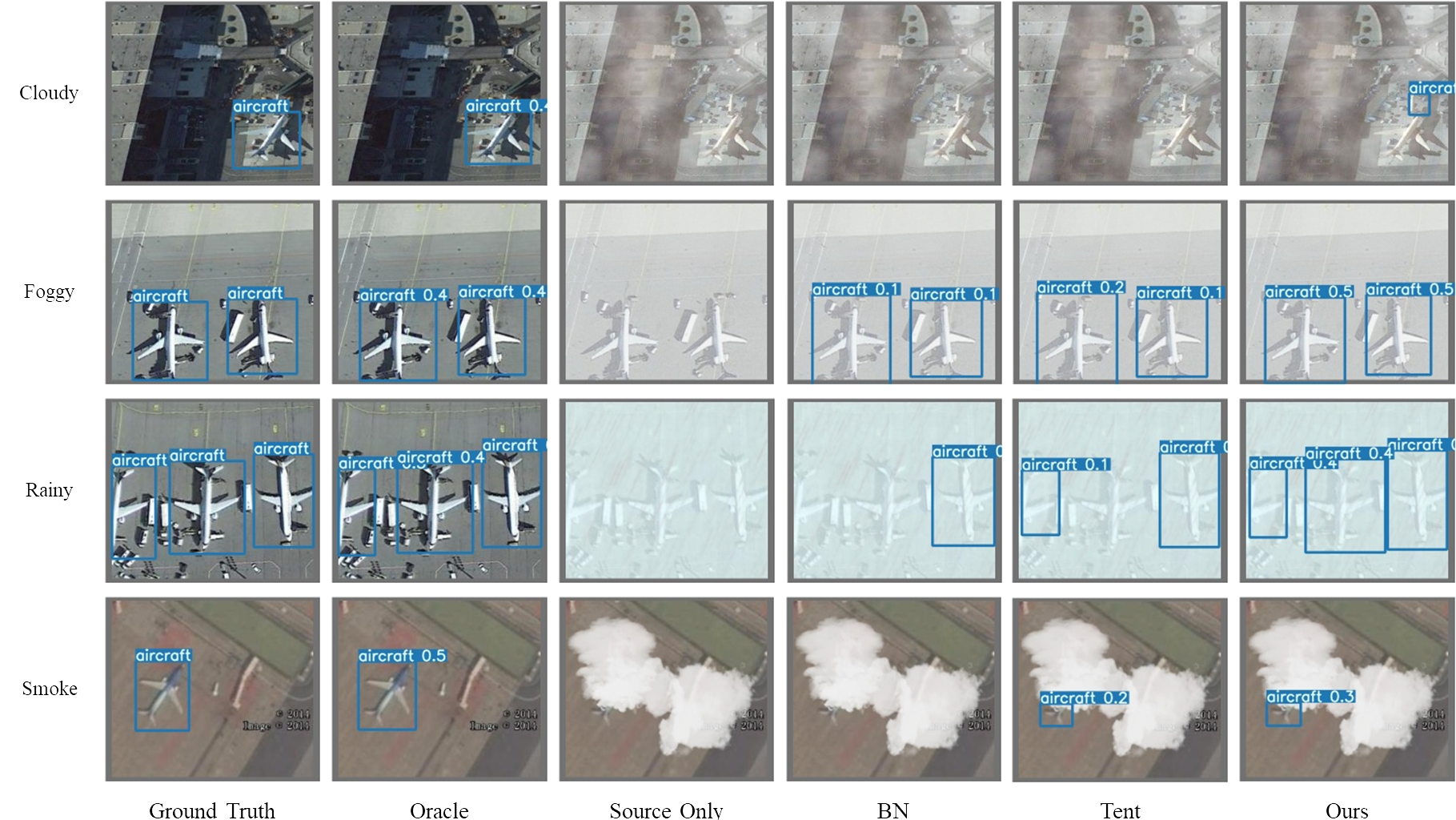}} 
\caption{Visualization of detection results on image samples from the RSOD dataset by the compared adaptation methods under different weather conditions. 
}
\label{detection}
\end{figure*}

The AdaptSegNet and ADVENT methods leverage adversarial training between the source and target domain data to enhance the generalization capability. On the two datasets, AdaptSegNet achieves 30.72\% and 32.16\% mIOU, while ADVENT reaches 32.59\% and 36.78\% mIOU, respectively. BN and Tent can further improve the adaptation performance to 36.7\% and 40.52\% on the DLRSD dataset, and to 42.81\% and 44.51\% on the WHDLD dataset, respectively. BN and Tent do not require access to the source domain data, highlighting their significant advantage in source-free adaptation scenarios. Our method shows the best adaptation performance, achieving mIoU of 41.28\% and 45.20\% on the DLRSD and WHDLD datasets, respectively, surpassing all other compared methods. 
Figs. \ref{segmentation} visualizes segmented results on samples from the WHDLD dataset by the proposed method in comparison with the source model.
The results show the impact of weather interference on the performance of the source model and demonstrate the effectiveness of the proposed SNN adaptation method.

\subsection{Results on the Detection Task}\label{detectionresult}

TABLE \ref{tab:detection} shows the detection performance on RSOD and LEVIR datasets. Mean mAP is used to evaluate the overall performance across different weather conditions. 
As shown in TABLE~\ref{tab:detection}, weather interference significantly impacts the performance of pre-trained detection models. On the RSOD dataset, the mean mAP degrades drastically from 82.71\% in the source domain to 19.09\% on the corrupted test data. Similarly, on the LEVIR dataset, the mean mAP decreases from 77.84\% to 18.73\%. This underscores the critical need for effective online adaptation of the pre-trained models when deployed in real-world scenarios to face diverse weather changes.

However, contrary to the results in previous classification and semantic segmentation experiments, Tent performs worse than BN in the detection task. 
On the RSOD dataset, Tent's mAP decreases by 0.71\% compared to BN, while on the LEVIR dataset, it decreases by 1.68\%. This suggests that calculating the entropy loss based on all the predicted instances without filtering cannot effectively guide the network adaptation. Our proposed strategy, however, utilizes a filtering function to identify high-confidence instances from the detection head outputs, enabling the model to effectively leverage reliable instances for adaptation. 
On the RSOD dataset, our method improves the mean mAP of the source model from 19.09\% to 47.05\%. Similarly, on the LEVIR dataset, our method improves the mean mAP from 18.73\% to 42.12\%. on both the datasets, our method outperforms BN and Tent, achieving the best adaptation performance.
Figs. \ref{detection} visualizes detected results on image samples from the RSOD dataset by the compared adaptation methods under different weather conditions.

\subsection{Ablation Study and Analysis}

A series of ablation experiments are conducted to  disclose the contributions of each component of the proposed method, including the online algorithm, the adaptive activation scaling scheme, the temperature smooth in (\ref{temperature}), and the confidence-based instance weighting approach. 

\begin{figure}[!t] 
\center{\includegraphics[width=8cm]{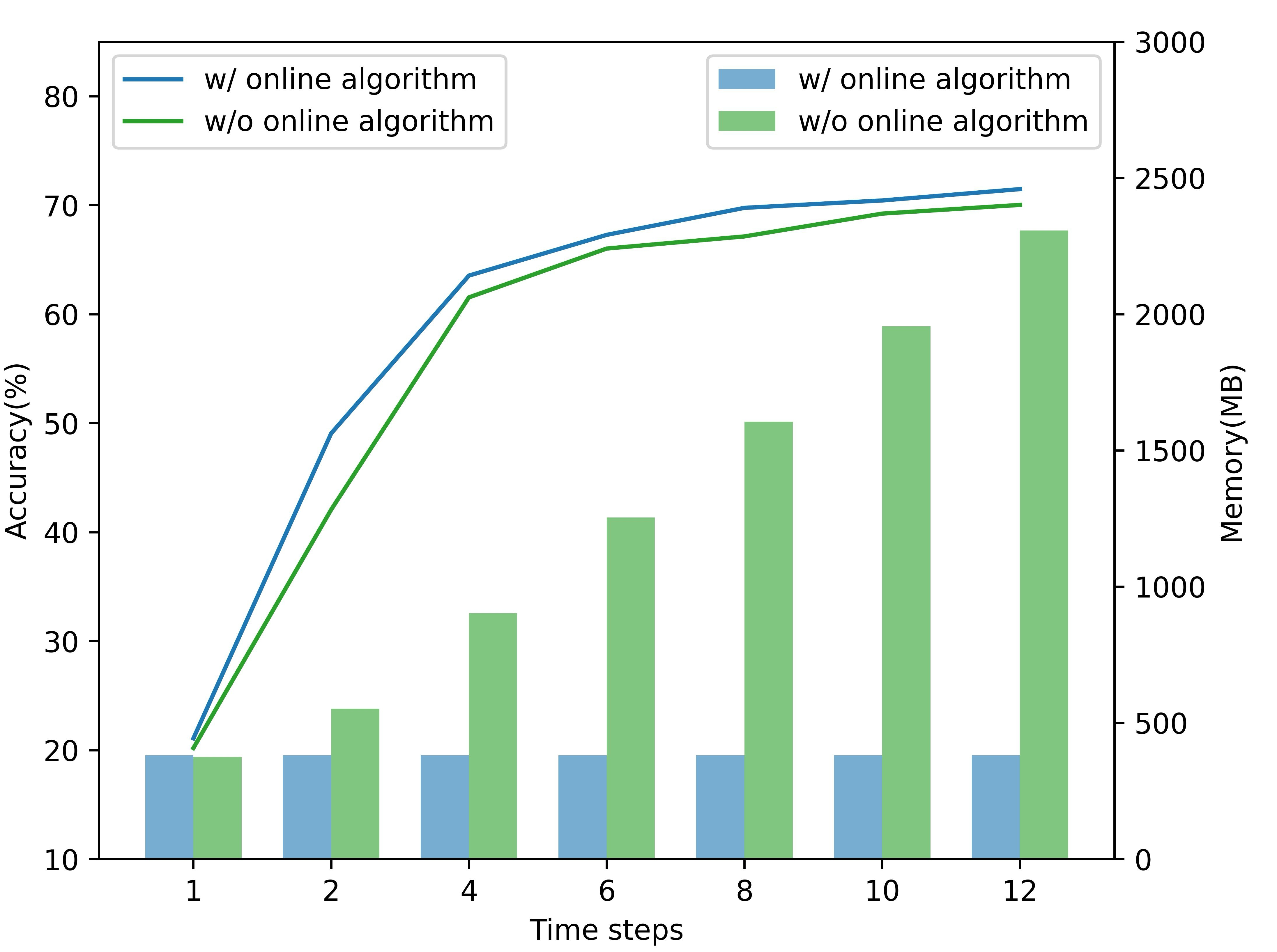}}
\caption{Accuracy and memory consumption of the SNN model during adaptation using BPTT or the online algorithm under different time steps.
}
\label{online_and_acc}
\end{figure}

First, we evaluate the effect of the online SNN adaptation algorithm introduced in Section \ref{online-alg} in comparison with the full BPTT algorithm introduced in Section \ref{IFBPTT}.
We conduct an one-epoch adaptation on the RSSCN7 classification dataset, monitoring both accuracy and memory consumption for different times-steps.
As shown in Figs.~\ref{online_and_acc}, the accuracy of both algorithms increase as the time-step increases, which is expected as SNNs encode information into binary temporal sequences. According to (\ref{bptt}), the standard full gradient backpropagation of SNN unrolls the model in the temporal domain, resulting in that the memory consumption increases linearly with the number of time-steps. In comparison, the memory consumption
of the online algorithm is constant, which would be more preferred in on-device adaptaion scenarios.

The second ablation study evaluates the effect of the proposed adaptive activation scaling (AAS) scheme. As introduced in Section (\ref{AAS}) and Fig. \ref{clip_histfig}, AAS adaptively controls the firing rate distribution of each spiking layer, making the SNN highly flexible during adaptation.

\begin{figure}[!t] 
\center{\includegraphics[width=8.5cm]{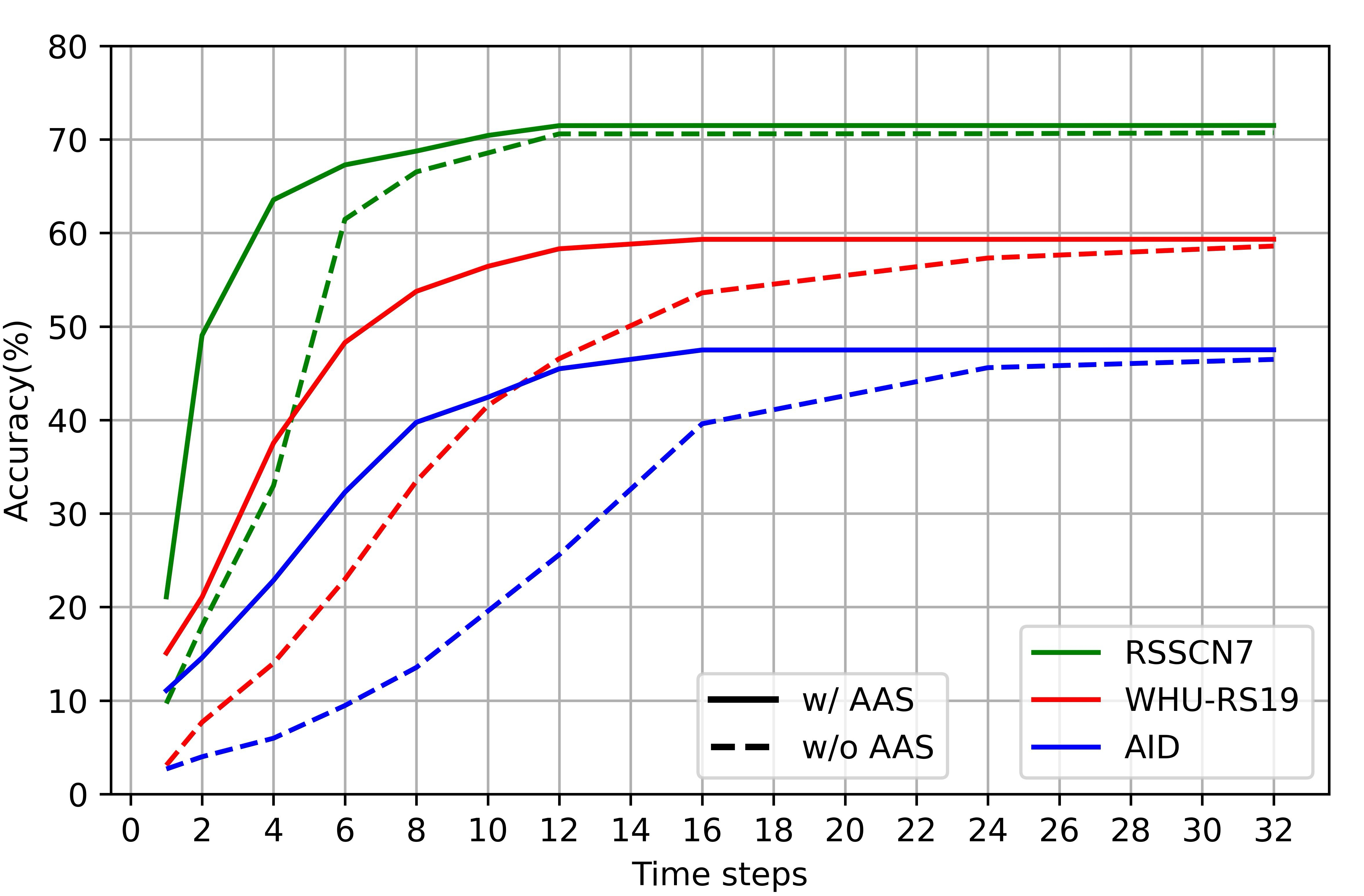}} 
\caption{Influence of the proposed adaptive activation scaling (AAS) scheme on the adaptation performance for varying time-steps on different datasets. VGG16 is used as the backbone network.}
\label{clip_and_acc}
\end{figure}

We conduct experiments on the RSSCN7, WHU-RS19, and AID datasets. Figs. \ref{clip_and_acc} shows the adaptation performance of the proposed method with and without the adaptive activation scaling scheme for different time-steps. It can be seen from Figs. \ref{clip_and_acc} that, with small time-steps, %e.g., $T<10$，
the performance of the SNN model is limited. 

However, under these conditions, using AAS can significantly enhance the model's adaptation performance. On the RSSCN7 dataset, the adaptation performance improves by over 30\% at 4 time-steps. On the WHU-RS19 and AID datasets, the performance increases by 26\% and 23\% respectively at 6 time-steps. 
On the whole, the proposed AAS scheme is especially effective at small time-steps. 
Using smaller time-steps can reduce and computational load and the processing latency. 
Thus, this advantage of the AAS scheme would be particularly preferred in applications,
such as on-platform processing with limited computing resources.

\begin{table}[!t]
\centering
\caption{Ablation study on the temperature parameter $\tau$ in calculating the instantaneous loss under different numbers of time steps.}
\label{ablationsmooth}
% \resizebox{0.8\columnwidth}{!}
{%
\begin{tabular}{@{}cccccccc@{}}
\toprule
\multirow{2}{*}{Dataset} & \multirow{2}{*}{Temp.} & \multicolumn{6}{c}{Time-steps} \\ \cmidrule(l){3-8} 
 &  & 2 & 4 & 8 & 12 & 16 & 32 \\ \midrule
\multirow{4}{*}{RSSCN7} & 1.0 & 30.78 & 45.78 & 47.96 & 49.58 & 51.78 & 54.82 \\
 & 2.0 & 35.28 & 50.37 & 53.04 & 56.29 & 58.92 & 60.75 \\
 & 4.0 & \textbf{49.06} & \textbf{63.54} & \textbf{68.76} & \textbf{71.48} & \textbf{71.5} & \textbf{71.53} \\
 & 8.0 & 40.86 & 55.89 & 59.24 & 62.08 & 63.38 & 64.93 \\ \midrule
\multirow{4}{*}{WHU-RS19} & 1.0 & 12.89 & 28.58 & 38.85 & 40.76 & 41.89 & 42.85 \\
 & 2.0 & 14.15 & 31.85 & 47.58 & 50.78 & 51.85 & 53.17 \\
 & 4.0 & \textbf{21.06} & \textbf{37.54} & \textbf{53.76} & \textbf{58.31} & \textbf{59.34} & \textbf{59.37} \\
 & 8.0 & 17.65 & 34.58 & 50.48 & 55.68 & 56.15 & 56.82 \\ \midrule
\multirow{4}{*}{AID} & 1.0 & 9.55 & 12.25 & 14.85 & 20.58 & 21.75 & 21.92 \\
 & 2.0 & 10.57 & 13.52 & 17.57 & 25.73 & 26.85 & 27.97 \\
 & 4.0 & \textbf{14.58} & \textbf{22.85} & \textbf{39.76} & \textbf{45.48} & \textbf{47.49} & \textbf{47.51} \\
 & 8.0 & 10.37 & 15.25 & 29.41 & 36.87 & 37.85 & 37.88 \\ \bottomrule
\end{tabular}%
}
\end{table}

The third ablation study examines the effect of the temperature smoothing calculating the instantaneous entropy loss in (\ref{instloss}) and (\ref{temperature}). 
TABLE~\ref{ablationsmooth} shows the results of proposed method with different values of the temperature parameters using a VGG16 backbone. Three datasets are considered, including the RSSCN7, WHU-RS19 and AID datasets. It can be seen that, without temperature
smoothing, the instantaneous entropy loss cannot achieve a satisfactory performance during adaptation. For example, on the RSSCN7 dataset with time-step $T=4$, without temperature smoothing it only yields an adaptation accuracy of 45.78\%, while using a temperature value $\tau=4$ it yields an adaptation accuracy of 63.54\%.
On the more complex AID dataset, the contribution of temperature smoothing is even more evident, e.g., the adaptation accuracy of the vanilla entropy loss only reaches about half that of the case with a temperature parameter $\tau=4$.

\begin{figure}[!t]%
    \centering
    \subfloat[Vanilla entropy loss without temperature smoothing]{
        \includegraphics[width=0.7\linewidth]{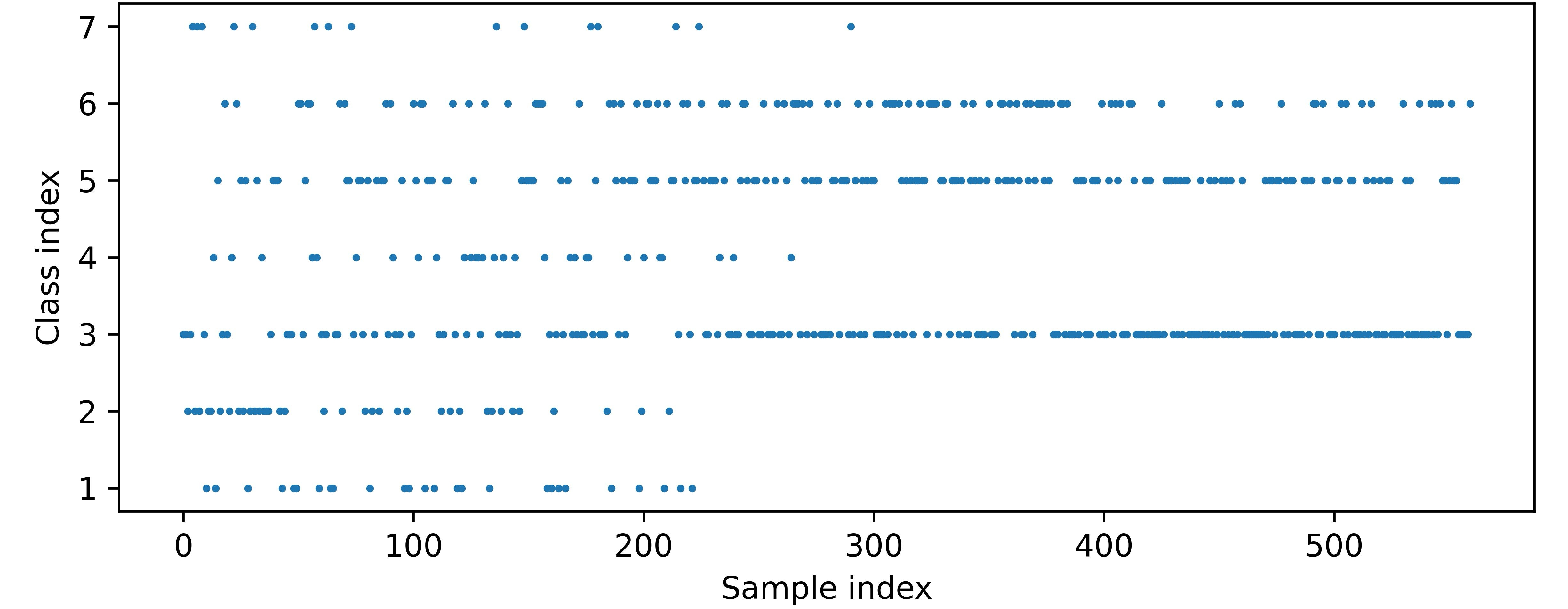}
        }\\
    \subfloat[Entropy loss with temperature smoothing]{
        \includegraphics[width=0.7\linewidth]{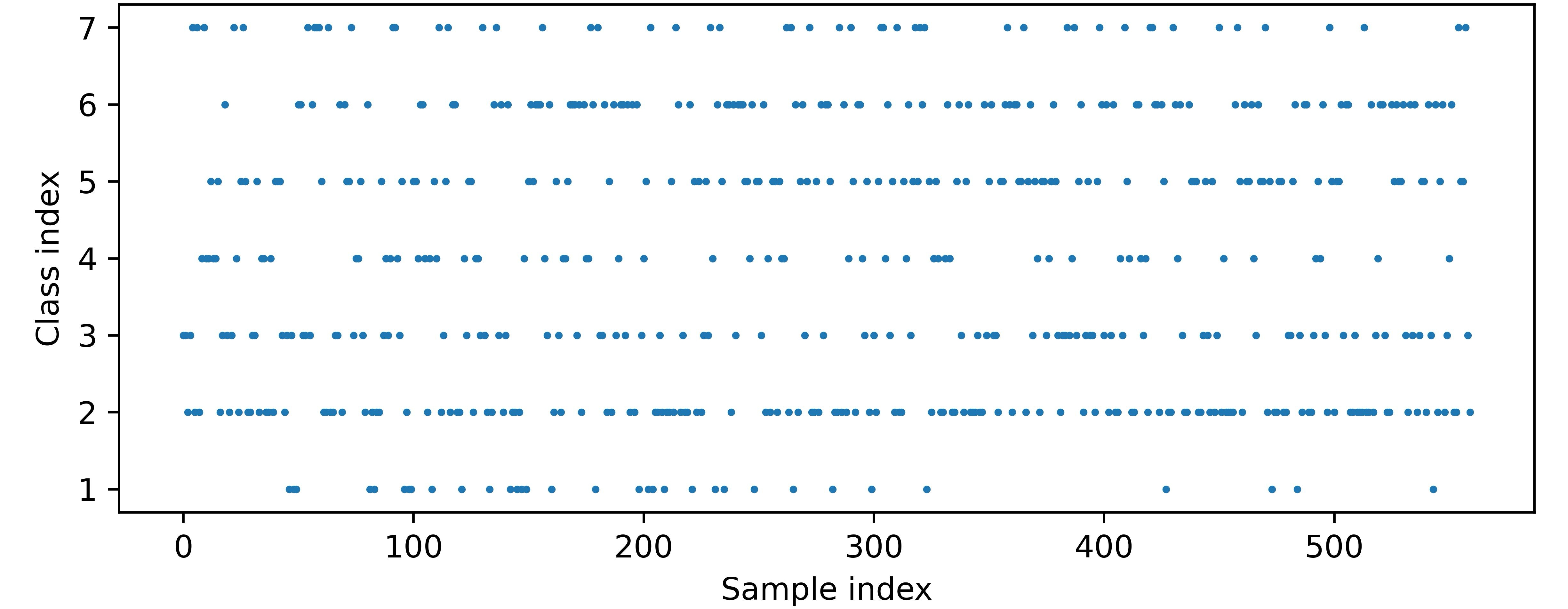}
        }
    \caption{Prediction distribution during adaptation of proposed SNN model. (a) Entropy loss with temperature smoothing ($\tau=4$). (b) Vanilla entropy loss without temperature smoothing.
    }
    \label{prediction_tem}
\end{figure}

To better analyze the effect of entropy temperature smoothing, 
we monitor the prediction distribution of the SNN model during one epoch adaptation on the RSSCN7 dataset. As shown in Figs.~\ref{prediction_tem}, when using appropriate temperature smoothing, the distribution of model predictions keeps uniform during adaptation as expected since the test samples have a uniform class distribution. In contrast, when using the vanilla entropy loss without temperature smoothing, the predictions tend to abnormally concentrate on a few classes, which is due to model collapse during adaptation. Basically, from to the physical meaning of entropy, the model can achieve the low entropy objective by constantly generating high-confidence predictions regardless the model inputs. This corresponds to a collapse issue of the entropy minimization loss.

\begin{table}[!t]
\centering
\caption{Ablation study on the confidence-based weighting approach with different boundary parameters $\tau_1$ and $\tau_2$.
``w/o CW" denotes the case without using the confidence based weighting approach.
}
\label{detection_ablation}
% \resizebox{\columnwidth}{!}
{
\begin{tabular}{clcccccc}
\hline
Method & \multicolumn{1}{c}{$\tau_1$} & $\tau_2$ & Cloudy & Foggy & Smoke & Rainy & Mean  mAP \\ \hline
BN & \multicolumn{2}{c}{-} & 39.45 & 45.24 & 47.34 & 48.82 & 45.21 \\
\hline
\multirow{4}{*}{Ours} & 0.1 & 0.9 & 40.87 & 46.53 & 48.79 & 48.98 & 46.29 \\
 & 0.2 & 0.8 & \textbf{41.75} & \textbf{46.75} & \textbf{49.85} & \textbf{49.83} & \textbf{47.05} \\
 & 0.3 & 0.7 & 39.76 & 46.04 & 48.42 & 49.15 & 45.84 \\
 & \multicolumn{2}{c}{w/o CW} & 39.12 & 45.15 & 46.17 & 47.56 & 44.50 \\ \hline
\end{tabular}
}
\end{table}

Lastly, we examine the effect of the proposed confidence-based weighting approach on the detection task.
As shown in  (\ref{weight}) and (\ref{detectloss}), the parameters $\tau_1$ and $\tau_2$ determine the confidence boundary for high-confidence selection. TABLE \ref{detection_ablation} shows the results of proposed method with different values of $\tau_1$ and $\tau_2$ on the RSOD dataset. It can be seen that, without using the weighting approach, that is indiscriminately calculating the entropy loss on all the predicted instances, the performance of the proposed adaptation method is even inferior to the simple BN method, with mean mAP 44.5\% versus 45.21\%.
 
As the filtering range gradually expands, more low-confidence instances are filtered out, leading to an improvement in adaptation performance. With $\tau_1=0.2$ and $\tau_2=0.8$, our method achieves the highest mean mAP of 47.05\%. 
However, beyond a certain point, the performance degrades due to the insufficient number of instances available for calculating the entropy loss. 
For example, with $\tau_1=0.1$ and $\tau_2=0.9$, the mAP decreases to 46.29\%. 
Overall, with appropriate values of $\tau_1$ and $\tau_2$, the proposed weighting approach can significantly improve the adaptation performance on the detection task.

%能效的分析
\subsection{Energy Efficiency Analysis}

As the main motivation of this work is to exploit the high energy efficiency
of SNN for on-device remote sensing processing, here we provide a comparison
on its energy consumption against ANN. 
We compare the estimated energy consumption of models running on hardware. 
For ANNs, the computation operations are dominated by floating-point multiply-accumulate computation (MAC), while for SNNs, it only involves accumulate computation (AC) because spikes are binary. 
For a fair comparison, we compare the energy consumption of processing a single image with the same architecture of the backbones. 

For ANN, we record the total number of floating-point operations (FLOPs). 
For SNN, we record the total number of synapse operations (SynOps), 
which counts the AC operations triggered by fired spikes.
We follow the work \cite{energy} on energy consumption analysis, using a 45nm chip as a reference. 
According to \cite{energy}, in a 32-bit floating-point MAC operation, the multiplication consumes 3.7pJ, the addition consumes 0.9pJ, and hence a single MAC operation consumes 4.6pJ. 
In contrast, a 32-bit integer AC operation consumes 0.1pJ. 
TABLE~\ref{tab:energy} provides the energy consumption analysis for the three backbone architectures used in this work for both  ANN  and SNN. 
It can be seen that SNN has a tremendous advantage over its ANN counterpart in energy efficiency. Across different backbone architectures, SNN can achieve about two orders of magnitude higher energy efficiency when deployed on neuromorphic hardware. 

\begin{table}[!t]
\centering
\caption{Comparison of energy consumption of ANN and SNN for different backbone architectures.}
\label{tab:energy}
{
%\resizebox{\columnwidth}{!}
{
\begin{tabular}{@{}ccccc@{}}%
\toprule
Model & Type & FLOPs (G) & SynOps (G) & Energy (J) \\ \midrule
\multirow{2}{*}{VGG16}
 & ANN & 42.39 & - & 1.70 \\
 & SNN ($T\!=\!4$) & - & 2.27 & 2.04E-2 \\ \midrule
\multirow{2}{*}{ResNet34} 
& ANN & 11.63 & - & 4.65E-1 \\
& SNN ($T\!=\!4$)& - & 0.56 & 5.08E-3 \\ \midrule
\multirow{2}{*}{FCN} 
 & ANN & 9.28 & - & 3.71E-1 \\
 & SNN ($T\!=\!8$)& - & 0.59 & 5.32E-3 \\ \bottomrule
 \end{tabular}%
}}
\end{table}

\section{Conclusion}

To fully leverage the advantages of brain-inspired computing for on-device remote sensing image processing, this work proposed an SNN based online adaptation framework, with application to the classification, semantic segmentation and detection tasks. 
We proposed an efficient unsupervised online adaptation method for SNN, in which
approximate BPTT algorithm is adopted to reduce the computational complexity.
Moreover, an adaptive activation scaling scheme is proposed to enhance the adaptation performance of SNN. Furthermore, a confidence-based instance
weighting scheme is designed to improve the adaptation
performance on the more challenging detection task.
Extensive experiments on seven remote sensing datasets, encompassing classification, semantic segmentation and detection tasks, demonstrated the effectiveness of the proposed method and its advantages over ANN methods. 

Considering the energy efficiency advantage of SNN, 
these results show a promising prospect 
of the deployment of SNN-based online adaptive learning methods on edge-devices with limited computing resources. %, e.g., on on-orbit satellites and UAVs.
In the future, with the development of neuromorphic computing hardware and dedicated chips for SNNs, the proposed method can be deployed on such dedicated hardware to efficiently run on edge-devices, such as on on-orbit satellites and high-altitude drones.
By utilizing adaptive SNN algorithms to process remote sensing images directly on edge-devices, the perception capabilities of remote sensing equipment can be greatly enhanced.

\bibliographystyle{unsrtnat}
\bibliography{mylib} 
\end{document}